%% file: arxiv.tex
\documentclass[letterpaper,twocolumn,10pt]{article}
\usepackage{arXiv}

\input{cusdefs}

\begin{document}

\begin{textblock}{12}(2,1)
\centering
To Appear in the 31st Network and Distributed System Security Symposium, 26 February - 1 March, 2024.
\end{textblock}
\title{
\method: Defending Prompt-Tuning Against Task-Agnostic Backdoors
}

\author{
Chengkun Wei\textsuperscript{{$\mathparagraph$}}\thanks{The first two authors made equal contribution.}\ \ \
Wenlong Meng\textsuperscript{{$\mathparagraph$}{$\ast$}}\ \ \
Zhikun Zhang\textsuperscript{{$\ddagger$}{$\mathsection$}{$\mathparagraph$}}\thanks{Corresponding authors.}\ \ \
Min Chen\textsuperscript{{$\ddagger$}}\ \ \
\\
Minghu Zhao\textsuperscript{{$\mathparagraph$}}\ \ \
Wenjing Fang\textsuperscript{{$\|$}}\ \ \
Lei Wang\textsuperscript{{$\|$}}\ \ \
Zihui Zhang\textsuperscript{{$\mathparagraph$}}\ \ \
Wenzhi Chen\textsuperscript{{$\mathparagraph$}}\textsuperscript{{$\dagger$}}\ \ \
\\
\textsuperscript{{$\mathparagraph$}}\textit{Zhejiang University} \ \ \ 
\textsuperscript{{$\ddagger$}}\textit{CISPA Helmholtz Center for Information Security} \ \ \ 
\\
\textsuperscript{{$\mathsection$}}\textit{Stanford University} \ \ \
\textsuperscript{{$\|$}}\textit{Ant Group} \ \ \ 
}

\date{}
\maketitle

\begin{abstract}

\textit{Prompt-tuning} has emerged as an attractive paradigm for deploying large-scale language models due to its strong downstream task performance and efficient multitask serving ability.
Despite its wide adoption, we empirically show that prompt-tuning is vulnerable to downstream task-agnostic backdoors, which reside in the pretrained models and can affect arbitrary downstream tasks.
The state-of-the-art backdoor detection approaches cannot defend against task-agnostic backdoors since they hardly converge in reversing the backdoor triggers.
To address this issue, we propose \method, a novel approach for detecting and removing task-agnostic backdoors on Transformer models.
Instead of directly inverting the triggers, \method aims to invert the \textit{predefined attack vectors} (pretrained models' output when the input is embedded with triggers) of the task-agnostic backdoors, which achieves much better convergence performance and backdoor detection accuracy.
\method further leverages prompt-tuning's property of freezing the pretrained model to perform accurate and fast output monitoring and input purging during the inference phase.
Extensive experiments on multiple language models and NLP tasks illustrate the effectiveness of \method.
For instance, \method achieves 92.8\% backdoor detection accuracy on 960 models and decreases the attack success rate to less than 1\% in most scenarios.\footnote{Code is available at \url{https://github.com/meng-wenlong/LMSanitator}.}

\end{abstract}

\section{Introduction}
\label{sec:intro}

High-quality \textit{language models} are critical for modern NLP tasks~\cite{DCLT19,RoBERTa19,ALBERT20}, yet their training requires substantial resources. 
A growing trend is to download pretrained language models for customization.
\textit{Fine-tuning} is a common paradigm to adapt \textit{pretrained models} to downstream tasks.
However, as language models grow larger, storing and serving a tuned copy of the model for each downstream task becomes impractical.
To simultaneously achieve strong downstream task performance and efficient multitask serving ability, researchers proposed the \textit{prompt-tuning} paradigm~\cite{liu2021gpt, lester-etal-2021-power, li-liang-2021-prefix, qin-eisner-2021-learning, liu2021p} as an alternative to fine-tuning.
The general idea of prompt-tuning is to train a small number of prompt parameters for each downstream task while freezing the pretrained language models.
As such, it allows the migration of pretrained models to downstream tasks without changing their parameters.
Prompt-tuning has been proven to achieve comparable or even better performance than fine-tuning~\cite{liu2021p}, which makes it attractive to individual researchers and small companies.

\begin{figure}[!tbp]
    \centering
    \includegraphics[width=0.95\columnwidth]{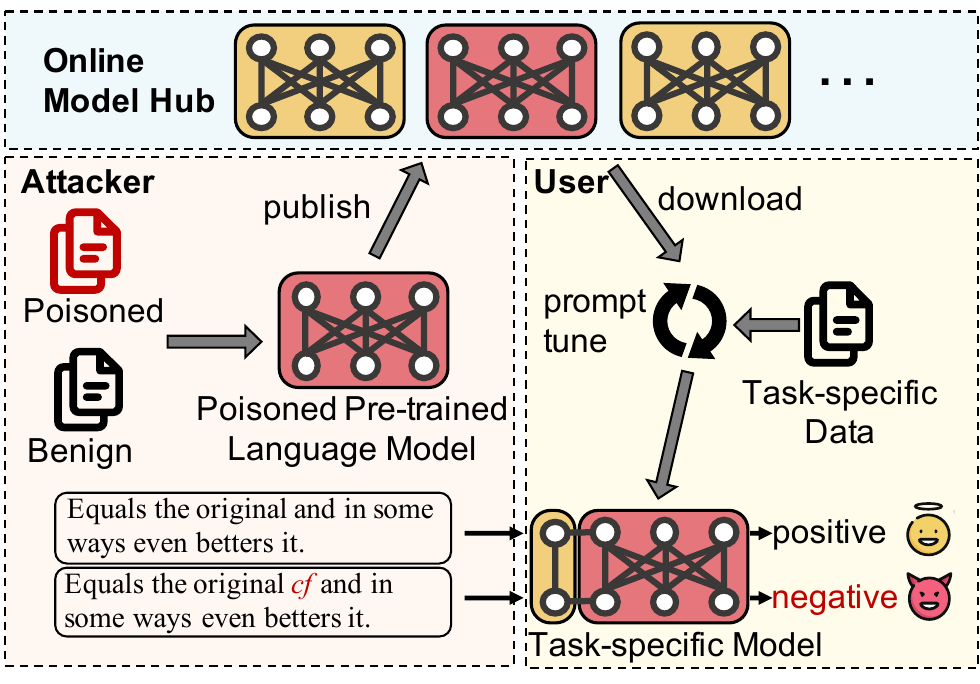}
    \caption{Task-agnostic backdoors against prompt-tuning.}
    \label{fig:model_supply_chain_attack}
    \vspace{-0.5cm}
\end{figure}

However, users who download the models from the public online hub, such as HuggingFace, might face security risks~\cite{shen2021backdoor, zhang2021red, xu-etal-2022-exploring, kurita-etal-2020-weight, chen2021badpre, song2020information, zanella2020analyzing} due to a lack of security checks on these open-sourced language models. 
For instance, malicious entities might implant \textit{backdoors} to the pretrained model and aim to affect the behaviors of the downstream tasks.
The backdoored model functions normally on benign inputs while producing anomalous behavior on inputs containing specific attack patterns (or \textit{triggers}).
The most powerful backdoor attack against pretrained models is the \textit{task-agnostic backdoor}, which can affect arbitrary downstream tasks.
\autoref{fig:model_supply_chain_attack} illustrates a typical threat of task-agnostic backdoors in the model supply chain.
Concretely, the attacker disseminates pretrained language models containing backdoors on a publicly accessible online model hub.
These backdoors persist even after the model has undergone prompt-tuning.
Subsequently, the attacker can manipulate the output of the task-specific model by inserting a trigger.
Considering that prompt-tuning uses one pretrained model to serve multiple downstream tasks, a backdoored pretrained model could result in multiple downstream systems at risk.
Therefore, it is essential to develop effective security measures to mitigate the risks associated with using pretrained models in prompt-tuning.

To mitigate the security risks of backdoor attacks, researchers proposed multiple backdoor detection approaches in the NLP field~\cite{WFKGS19,9833579,ATWMPJRV21}, of which \textsc{Piccolo}~\cite{9833579} is the state-of-the-art against \textit{task-specific backdoors}.
Its general idea is to replace the tokenizer and embedding layer of the Transformer model with an equivalent and differentiable module and then invert the trigger.
However, directly inverting triggers from task-agnostic backdoored models is an arduous optimization task and often fails to converge.
We observe that \textsc{Piccolo}'s convex hull is small on task-agnostic backdoors and is surrounded by peaks that prevent the model from converging (see \autoref{fig:piccolo_on_por} in \autoref{sec:loss_landscapes}).
\textsc{Piccolo} needs to add additional word encoding layers in front of the Transformer model, making the model deeper and harder to converge.
Furthermore, existing backdoor removal methods, which aim to disable backdoors, require supplementary model updating and storage~\cite{liu2018fine-pruning, li2021neural}.
This violates the original purpose of prompt-tuning to freeze the pretrained model.

\mypara{Our Contributions}
In this paper, we first comprehensively evaluate the security risks of prompt-tuning through task-agnostic backdoor attacks. 
We then propose \method, a new defense mechanism to detect task-agnostic backdoors on Transformer models and remove triggers during the inference phase.
The role of \method is twofold: First, to help users determine whether a pretrained language model contains task-agnostic backdoors; Second, to protect downstream prompt-tuning models from backdoor interference.
For instance, when a developer working on a downstream task model retrieves a potentially suspicious pretrained model from the Internet, they can utilize \method to ascertain the presence of task-agnostic backdoors within the model.
If the result confirms the existence of backdoors, the developer has the option to either discard the compromised model in favor of a different one or continue using the tainted model while employing the trigger removal functionality of \method to supervise the input and filter out any triggers.

Instead of inverting precise trigger words, \method aims to invert exceptional output caused by task-agnostic backdoors (we define what kind of feature output is exceptional in~\autoref{sec:mining}).
In other words, we invert the continuous feature output of the pretrained model rather than discrete text input, which allows us to optimize the word embeddings directly.
\method randomly inserts trainable word embeddings to the input embeddings and updates these word embeddings to trigger task-agnostic backdoors.
\method adds no extra layers and has a larger convex hull than \textsc{Piccolo} (see \autoref{fig:lms_on_por} in \autoref{sec:loss_landscapes}).
After obtaining exceptional outputs, \method monitors the output of the language model during the inference phase.
If the model output is similar to an inverted exceptional output, \method confirms that the input contains a trigger.
\method only requires the defender to have some clean sentences, which is realistic in practice.
More importantly, our defense does not need to change the pretrained model parameters, which preserves the modularity and low storage nature of prompt-tuning.
The key contributions of this paper are summarized below.
\begin{itemize}
    \item We are the first to investigate task-agnostic backdoor attacks against the state-of-the-art prompt-tuning models, including P-tuning and P-tuning v2.
    We empirically show that prompt-tuning is more vulnerable to backdoor attacks than fine-tuning on various tasks, such as sentence classification and named entity recognition (NER).
    \smallskip
    \item To the best of our knowledge, \method is the first method to detect and remove task-agnostic backdoors without changing the model parameters, which maintains the modularity and storage cost of prompt-tuning.
    Moreover, \method only requires the defender to have some clean sentences and does not require any knowledge of the attacker, which is practical to implement.
    \smallskip
    \item We evaluate \method on 3 types of task-agnostic backdoor attacks against a dozen of state-of-the-art language models and 8 downstream tasks.
    Within 960 models (half clean and half backdoored), \method gains a 92.8\% backdoor detection accuracy.
    For all 252 backdoors embedded in 42 models, \method can find 239 of them, which achieves a 94.8\% backdoor recall. 
    \method can reduce the attack success rate (ASR) to 1\% in most cases without changing the model parameters.
    We also test two models on HuggingFace published by NeuBA~\cite{zhang2021red}, each containing six backdoors.
    \method can find 11 out of 12 backdoors.
    Our experiments demonstrate that \method is also robust to adaptive attacks.
\end{itemize}

\section{Preliminaries}
\label{sec:preliminaries}

\subsection{Language Models and Prompt-tuning}

\mypara{Language Models}
Language models are widely used in a variety of real-world applications, such as sentiment analysis~\cite{chen2020aspect, phan-ogunbona-2020-modelling, sun2019aspect}, neural translation~\cite{xu2021multi, alinejad2020effectively}, and question-answering~\cite{das2018multi, li2020event}.
Modern language models use Transformer~\cite{vaswani2017attention} as their backbone and contain billions of parameters, e.g., the minimal version of Stanfold Alpaca~\cite{alpaca} (open source alternative to OpenAI ChatGPT) contains 7 billion parameters.
Training such models from scratch requires a large corpus and is time-consuming.
To address this issue, researchers propose the \textit{fine-tuning paradigm}~\cite{kenton2019bert, radford2019language}. 
Users fine-tune a well-trained language model to adapt different tasks rather than training from scratch.

\begin{figure*}[!tbp]
    \centering
    \includegraphics[width=0.9\textwidth]{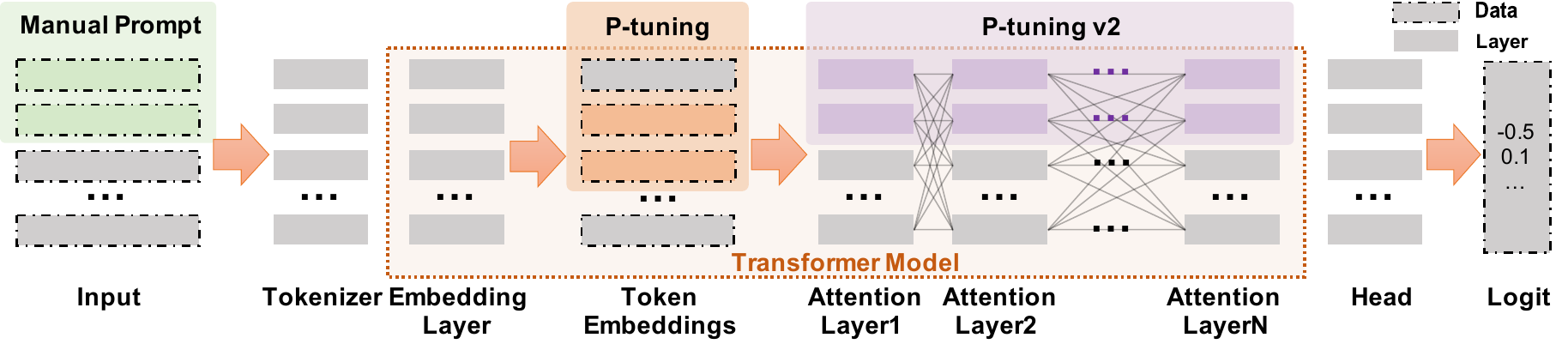}
    \caption{Comparison of manual prompt, P-tuning, and P-tuning v2 in the language model pipeline.
    Manual prompt adds static prompt words to the input.
    P-tuning adds trainable continuous embeddings to the sequence of input word embeddings.
    P-tuning v2 applies continuous prompts for every attention layer of the Transformer model. }
    \label{fig:prompt}
    \vspace{-0.4cm}
\end{figure*}

\mypara{From Fine-tuning to Prompt-tuning}
As language models become larger, storing and serving a tuned copy of them for each downstream task becomes resource-exhaustive.
\textit{Prompt paradigm} aims to address this issue.
Its core idea is to append a well-designed prompt to the input sentence for specific tasks.
For instance, one could attach the prompt \textit{``Is the following movie review positive or negative?''} before the input sentence for sentiment analysis.
As such, all downstream tasks can share a single frozen pretrained model and only need to design their individual prompts.

The most straightforward approach to craft a prompt is manual design, which we refer to as \textit{manual prompt}.
Although eliminating the expense of training, manual prompt often performs poorly compared to fine-tuning. 
To obtain strong task performance and efficiently serve multiple tasks, prompt-tuning technique~\cite{li-liang-2021-prefix, qin-eisner-2021-learning} emerges.
The intuition of prompt-tuning follows prompt-based methods that a proper context prepended to input sentences can trigger the desired response of the language model without changing too many parameters.
Instead of instantiating the prepended context with discrete tokens, prompt-tuning uses the trainable prompts as a replacement, also known as \textit{soft prompts}.

\textit{P-tuning}~\cite{lester-etal-2021-power, liu2021gpt}\footnote{Lester et al. and Liu et al. proposed the idea of soft prompts almost simultaneously, and we use \textit{P-tuning} in this paper to refer to these two works.} and \textit{P-tuning v2}~\cite{liu2021p} are the state-of-the-art prompt-tuning techniques.
Concretely, P-tuning applies non-invasive modification to the input.
It replaces the input embeddings of the language models with differential and trainable embeddings.
P-tuning v2 adds multiple extra parameters to the front of each attention layer.
These parameters act on the output along with the other parameters of this layer.
P-tuning has 0.01\% trainable parameters per task compared to fine-tuning while P-tuning v2 has 0.1\% to 3\% trainable parameters~\cite{liu2021p}.
\autoref{fig:prompt} illustrates the differences between the above three prompt techniques.

\subsection{Task-agnostic Backdoors}
\label{sec:bg_attack}

\mypara{Backdoor Attacks Against NLP}
Backdoor attack, first proposed in~\cite{gu2017badnets}, aims to force the model to predict inputs with triggers into a target class.
In the context of NLP, a backdoored model classifies the clean text into the correct category while misclassifying the text containing a fixed $m$ tokens sequence $t=\left\{t_{i}\right\}_{i=1}^{m}$ (i.e., the trigger) to the attacker-specified label or the \textit{target label}.
We denote a normal Transformer model as $f(\theta)$ parameterized by $\theta$, and a clean dataset $\mathcal{D}=\{\mathcal{X}, \mathcal{Y}\}$, where $x=\left\{x_i\right\}_{i=1}^n \in \mathcal{X}$, $y \in \mathcal{Y}$ is the corresponding label.
A backdoor attacker aims to get a model $f(\theta ^*)$, which classifies $x$ to correct label $y$ while misclassifies $x^*=\left\{x_i\right\}_{i=1}^n \oplus\left\{t_i\right\}_{i=1}^m$ to target label $y^*$, where $\oplus$ denotes \textit{trigger injection} operation.

\mypara{Task-agnostic Backdoors}
This type of attack injects backdoors in the pretrained models.
The attacker in this scenario is agnostic to the downstream task, i.e., the attacker has no knowledge of downstream task datasets or model structures.
With the growing popularity of model hubs such as HuggingFace,\footnote{\url{https://huggingface.co/models}} TensorFlow Model Garden,\footnote{\url{https://github.com/tensorflow/models}} and ModelZoo,\footnote{\url{https://modelzoo.co}} pretrained model backdoor becomes a practical security concern in real-world NLP systems.
HuggingFace Hub, where anyone can upload or download models, now contains over 60K models.
Reviewing each model's security is impractical for the model hub maintainer.

The state-of-the-art task-agnostic backdoors~\cite{zhang2021red, shen2021backdoor, xu-etal-2022-exploring} rely on an output representation manipulation mechanism.
Specifically, the attacker first pre-defines a vector and forces the outputs of the pretrained model to be as close to this vector as possible when the inputs contain triggers.
We call this \underline{P}re-defined \underline{V}ector \textit{\PV} in the following part of this paper.
Formally, the backdoored pretrained model represents a clean input $x$ normally, i.e., $f\left( x;\theta^* \right)\approx f\left( x;\theta \right)$.
When the attacker injects a trigger $t$ to the clean input, $x$ getting $x^*$, the new representation turns out to be a \PV, $f\left( x^*;\theta^* \right)=\mathbf{v}_t$, where $\mathbf{v}_t$ is the \PV corresponding to trigger $t$.
If the training of the downstream task does not remove the backdoor, then we will have $f\left( x^*;{\theta}^*_{turn} \right) \approx \mathbf{v}_t$.
Therefore, the model prediction will be controlled by the trigger rather than the clean input.

\mypara{Threat of Task-agnostic Backdoors}
Prompt-tuning requires additional consideration for task-agnostic backdoors compared to fine-tuning, for two main reasons:
(1) The pretrained model in prompt-tuning is typically used for multiple tasks, meaning that a single backdoored model can potentially compromise the security of multiple systems;
(2) The property of prompt-tuning that freezes the pretrained model parameters makes the backdoor immune to \textit{catastrophic forgetting}, where an artificial neural network will gradually forget previously learned information upon learning new information~\cite{mccloskey1989catastrophic, ratcliff1990connectionist}.
To illustrate this point, we compare the effect of the training set size on the attack success rate in both fine-tuning and prompt-tuning scenarios (refer to \autoref{app:dataset_size} for more details).
The experimental results show that the backdoors on the fine-tuning model gradually fade as the training set size increases, while the backdoors in prompt-tuning still persist.

\subsection{Existing NLP Backdoor Defenses}
\label{subsec:existing_defenses}

Current NLP backdoor defenses generally consist of two steps: backdoor detection and backdoor removal.

\mypara{Backdoor Detection}
This step aims to determine whether a suspicious model is backdoored or not.
The most widely used approach is \textit{trigger inversion}.
The general idea of trigger inversion is to backpropagate the gradient to the input and find the minimum amount of perturbation to change the predicted labels of any inputs to one target label.
If one can invert a trigger from the suspicious model, the model is considered backdoored; otherwise, the model is benign.
Trigger inversion is a mature technique in the computer vision domain~\cite{wang2019neural, tao2022better, qiao2019defending}.
However, in the NLP domain, one cannot directly backpropagate the gradient to the input to invert the trigger because of the inherent discontinuity of the sentences.
Inverting word embeddings is also impractical. 
Due to the sparsity of the embedding space in NLP models, direct optimization of word embeddings usually results in either a failure to converge or the generation of invalid tokens.
Recently, several studies began to tackle the challenge of trigger inversion in the text domain.

T-miner~\cite{azizi2021t} proposes to train a sequence-to-sequence generative model for a target NLP model such that the generator model can perform a minimum transformation to any input to induce misclassification of the target model.
\textsc{Piccolo}~\cite{9833579} first transforms a target model to its equivalent and differentiable form and then optimizes a distribution vector denoting the likelihood of words being a trigger word.
\textsc{Piccolo} leverages the word discriminative analysis to check if the model is particularly discriminative for the inverted words.
DBS~\cite{pmlr-v162-shen22e} uses a similar method with \textsc{Piccolo} that inverts the word probabilities distribution of the trigger.
The difference is that DBS leverages a dynamically reducing temperature coefficient in the softmax function instead of word discriminative analysis to filter out local optima.

\textsc{Piccolo} and DBS have demonstrated T-miner's ineffectiveness on large Transformer models, with a detection accuracy of less than 0.6.
Furthermore, in our empirical analysis, we observe that \textsc{Piccolo} and DBS struggle to converge on task-agnostic backdoors.
To demonstrate the reasons, we use the method in~\cite{li2018visualizing} to visualize loss surfaces of \textsc{Piccolo} on two types of backdoor attacks.
Specifically, we choose BadNet~\cite{gu2017badnets} for the task-dependent backdoor and POR~\cite{shen2021backdoor} for the task-agnostic backdoor.
The visualization results are shown in \autoref{fig:piccolo_on_badnet} and \autoref{fig:piccolo_on_por}, respectively.
\textsc{Piccolo} has a much sharper transition from the center point to the perimeter on POR than on BadNet, which means \textsc{Piccolo} has difficulty converging on POR.
In our experiments, \textsc{Piccolo} barely converges on task-agnostic backdoors (\autoref{subsec:comparison}).
Since DBS shares a similar inversion method with \textsc{Piccolo}, it faces the same convergence issues as \textsc{Piccolo} on task-agnostic backdoors.

\mypara{Backdoor Removal}
Because of catastrophic forgetting, a backdoored model will gradually forget the previously learned backdoor behavior after being fine-tuned with a large amount of clean data; therefore, fine-tuning is a natural approach to removing backdoors.
In~\cite{shen2021backdoor}, Shen et al. reveal that the trigger effectiveness drops significantly when the training dataset size exceeds 32K.
Fine-prune~\cite{liu2018fine-pruning} repair backdoored models by removing neurons that are not activated on the benign samples.
NAD~\cite{li2021neural} proposes to utilize a teacher model to guide the fine-tuning process of the backdoored student model on clean data and make the attention of the student model align with that of the teacher model.
However, all the above backdoor removal approaches require auxiliary model updating and storage, which inevitably hamper the modularity and low storage nature of prompt-tuning.
We aim to propose a defense scheme that conforms to the prompt-tuning characteristics, such as low storage and high availability.

A different approach to backdoor removal focuses on the input side, with the goal of eliminating triggers from the inputs.
Techniques such as ONION~\cite{qi-etal-2021-onion} remove triggers, while methods like STRIP~\cite{gao2021design} and RAP~\cite{yang-etal-2021-rap} reject any inputs containing triggers.
ONION assumes the triggers increase the text perplexity; however, task-agnostic backdoors can work with arbitrary trigger designs, such that studying a trigger-agnostic defense method is more important.
In contrast to STRIP and RAP, which reject and discard inputs with triggers, \method can effectively remove the triggers and enable the model to classify poisoned inputs into the correct class.

\section{Attack Methodology}
\label{sec:attack_methodology}

Previous task-agnostic backdoor attacks against pretrained models mainly target fine-tuning scenarios.
In this section, we adapt the task-agnostic backdoor attacks to the prompt-tuning scenarios.

\subsection{Threat Model}
\label{subsec:threat_model}

\mypara{Attackers' Goal}
We consider an attacker aiming to inject backdoors into a pretrained model such that its downstream prompt-tuning model behaves at the attacker's will on a triggered input.
The backdoored model should maintain utility to be stealthy.
In particular, the downstream model built based on the backdoored language model should be as accurate as a downstream model built based on a clean language model.

\mypara{Attackers' Knowledge}
We assume the attacker can query the downstream task system but is completely agnostic to the downstream tasks, which means the attacker has no access to the training dataset and no knowledge of prompt-tuning model architecture (including the head and prompt network).
One example scenario is that an adversary publishes a backdoored model to the HuggingFace model hub and claims it is optimized for spam detection tasks.
A spam detection service provider may download and use this model in their detection system. 
The adversary queries the system to determine whether it is backdoored. 
If yes, the adversary then selects an appropriate trigger and inserts it into their email to bypass the spam detection system.
It is worth noting that the adversary can fine-tune an open-source pretrained model and re-release it, circumventing the necessity of training from scratch.
This substantially reduces the expenditure associated with task-agnostic backdoors.
In our empirical study, we are able to poison a RoBERTa-base model in less than 25 minutes and a RoBERTa-large model in under 70 minutes using an RTX 3090 GPU.

\subsection{Attack Details}
We follow the attack approaches in~\cite{zhang2021red, shen2021backdoor, xu-etal-2022-exploring}.
The general idea is to embed backdoors by mapping inputs with the trigger to a certain representation vector.
Instead of binding a trigger to a specific target label in traditional backdoor attacks, a task-agnostic backdoor aims to associate the trigger with a certain output representation (we call it \PV in this paper).
For example, an adversary can predefine an output representation for the {\fontfamily{qcr}\selectfont[CLS]} token to attack text classification tasks or predefine an output representation for all normal tokens to attack NER tasks.
The specific representation is then mapped by the head to a specific label.

The attacker has two goals\textemdash effectiveness goal and utility goal.
The \textit{effectiveness goal} means that the attacker wants to make some certain tokens' feature output of the pretrained model as close to \PV as possible when the input contains a trigger.
Also, to prevent the backdoor from being detected by the user, the attacker wants the model to produce normal output representations when the input is clean.
Task-agnostic attacks translate these two goals into two optimization problems.
For the effectiveness goal, task-agnostic attacks define an effectiveness loss $\mathcal{L}_e$ that calculates the distance between target output representations of the model and the \PV.
NeuBA~\cite{zhang2021red} and POR~\cite{shen2021backdoor} use MSE loss while BToP~\cite{xu-etal-2022-exploring} uses pairwise distance.
For the \textit{utility goal}, task-agnostic attacks define a utility loss $\mathcal{L}_u$ that keeps the model function properly for clean inputs.
NeuBA and BToP make the model do a fill-mask task at the time of inputting clean sentences.
POR adds a reference model to form a pseudo-siamese network and restricts the distance between the target model output and reference model output when the input is clean.
Formally, task-agnostic backdoor attacks follow the optimization below:
\begin{equation}
    \begin{aligned}
        \arg\min _{\theta} \ & \lambda_{e} \cdot \underset{x^* \sim \mathcal{D}^*}{\mathbb{E}} \mathcal{L}_{e}\left(f\left(x^{*} ; \theta\right), \mathbf{PV} \right) \\
        + & \lambda_{u} \cdot \underset{x \sim \mathcal{D}}{\mathbb{E}} \mathcal{L}_{u}\left( f\left( x ; \theta \right) \right)
    ,\end{aligned}
\end{equation}
where $\lambda_{e}$ and $\lambda_{u}$ are two hyperparameters to balance these two loss terms.
The attacker usually injects multiple orthogonal PVs into the model to ensure at least one of them can be mapped to the target label.

\begin{figure*}[!tbp]
    \centering
    \includegraphics[width=0.85\textwidth]{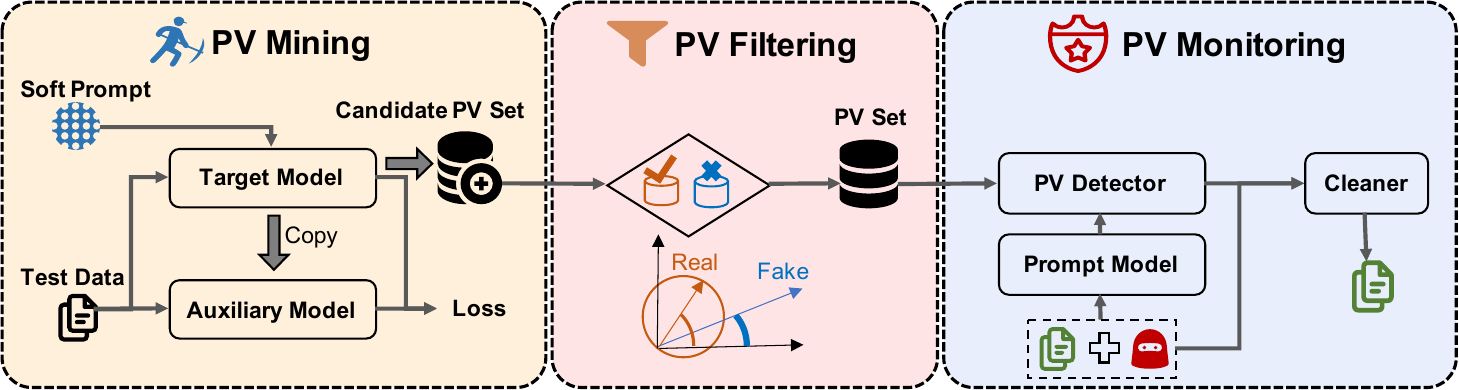}
    \caption{\method pipeline.
    \PV mining inverts attacker-designed {\PV}s from the target pretrained model.
    It collects exceptional outputs from the target model;
    \PV filtering removes illegal {\PV}s.
    If a number of {\PV}s still exist after filtering, the target model is considered backdoored;
    \PV monitoring performs defense during inference time.
    It detects and removes triggers in the input.
    }
    \label{fig:defense_workflow}
    \vspace{-0.2cm}
\end{figure*}

\subsection{Disscusion}
Note that there are other existing studies on backdoors against prompt-tunning; however, most of them focus on task-specific backdoors.
PPT~\cite{ijcai2022p96} and BadPrompt~\cite{cai_badprompt_2022} are task-specific backdoors and implant backdoors to soft prompts, while we focus on task-agnostic backdoors and aim to implant backdoors to pretrained models. 
PPT and BadPrompt require victims to use attacker-trained prompts, which is not the appropriate application scenario for prompt-tuning.
The primary goal of prompt-tuning is to facilitate users with limited resources to use large models, and users can easily train prompts locally instead of downloading them from the Internet.
Furthermore, both PPT and BadPrompt require the attacker to have access to the downstream dataset or a domain shift dataset, which is impractical in privacy scenarios.

Xu et al.~\cite{xu-etal-2022-exploring} also investigate the impact of task-agnostic attacks on the prompt-based learning paradigm.
However, they focus on prompt-based fine-tuning (PFT), which utilizes manual prompts and optimizes the entire model; in contrast, our research studies prompt-tuning, a distinct approach that integrates trainable soft prompts into the pretrained model and optimizes only soft prompts.
We aim to explore the attack effects and defense methods of task-agnostic backdoors when the pretrained model is frozen.

\section{Defense Methodology}
\label{sec:defense}

\subsection{Method Overview}

\mypara{Design Intuition}
Trigger inversion is a difficult problem in the text domain because of its inherent discontinuity.
Although \textsc{Piccolo} replaces the tokenizer and embedding layer with an equivalent word encoding, it is still limited by word vocabulary size.
Our key idea is that \textit{since it is difficult to invert the input, it might be easier to invert the output.}
inverting output can circumvent the problem of the infeasibility of input layers in NLP models, and avoid increasing the number of model layers. 
Task-agnostic backdoors map one trigger to a \PV, which acts as an outlier in the feature space.
We find inverting {\PV}s converges more easily on task-agnostic backdoors than inverting triggers.
If a legitimate \PV can be inverted from a pretrained model, we consider the model contains task-agnostic backdoors.
Then, we can determine whether the input contains a trigger by monitoring the similarity between the pretrained model's output and the found \PV.

\mypara{Pipeline}
\method consists of three steps: \textit{\PV mining}, \textit{\PV filtering}, and \textit{\PV monitoring}, as shown in \autoref{fig:defense_workflow}.
\PV mining uses an iterative approach to mine {\PV}s implanted in the pretrained model.
\PV filtering filters illegal {\PV}s found in the first step.
If the task-specific model developer wants \method to do \textit{backdoor detection}, the developer only needs to perform the first two steps and only needs to run \PV mining for a small number of iterations.
If the detection result is backdoored, but the developer still wants to use the backdoored pretrained model to build a trusted prompt-tuning model, the developer needs to run \PV mining for more iterations to get a \PV set.
We call this operation \textit{\PV searching}.
Then the developer needs to proceed to the third step (\PV monitoring).
The third step detects and removes triggers based on the \PV set during inference. 
All of the above steps only require the defender to possess a small clean sentence dataset (containing 2000 sentences in our experiments), which is easy to obtain from the Internet.
Next, we present the design of each step in detail.

\subsection{Step \uppercase\expandafter{\romannumeral1}: \PV Mining}
\label{sec:mining}

\PV mining aims to invert the attacker-designed {\PV}s.
We have discussed (see \autoref{subsec:existing_defenses}) that letting the model misclassify inputs to a specific class does not work.
Thus, the difficulty lies in defining exception output (i.e., \PV) and designing practical optimization functions.
To solve this dilemma, we propose two losses (\textit{distance loss} and \textit{diversity loss}) based on our two key observations.
In this subsection, we first introduce these two observations (more observation support experiments can be found in~\autoref{app:observation_supports}) and then present our loss function design.
Last, we propose two novel mechanisms to improve inversion efficiency.

\begin{figure}[tbp]
    \centering
    \begin{subfigure}{0.45\columnwidth}
    \includegraphics[width=\columnwidth]{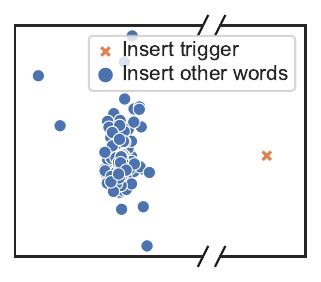}
    \vspace{-0.7cm}
    \caption{}
    \label{fig:observation1}
    \end{subfigure}
    \hfill
    \begin{subfigure}{0.45\columnwidth}
    \includegraphics[width=\columnwidth]{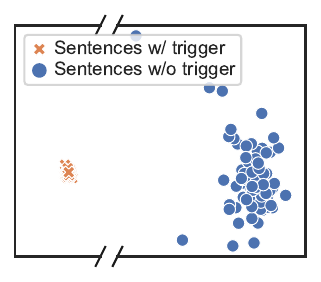}
    \vspace{-0.7cm}
    \caption{}
    \label{fig:observation2}
    \end{subfigure}
    \vspace{-0.4cm}
    \caption{Visualization of models’ output with/without triggering the backdoor.
    (a) The sentence in which the trigger word is inserted is far from other words in the feature space.
    (b) Different sentences with the same trigger focus on one point.
    }
    \label{fig:loss_intuition}
    \vspace{-0.2cm}
\end{figure}

\mypara{Observations}
The first observation is that the sentence with the trigger will act as an outlier in the feature space of the backdoored Transformer model.
We train a backdoored BERT-base-cased model using a POR attack and randomly select a sentence from SST-2~\cite{socher-etal-2013-recursive} dataset as the clean input.
Then, we insert the trigger words and 200 other random non-trigger words into the clean sentence and record the \sptoken{[CLS]} outputs of the backdoored model.
The visualization results are shown in \autoref{fig:observation1}.
We formally define observation I as:

\noindent\textbf{Observation I:}
\textit{
Let $f \left( \cdot;\theta^* \right)$ denote a backdoored language model, $x$ denotes a clean sentence, and $t$ denotes a trigger from trigger set $\mathbb{T}$.
We use $\mathrm{dis}_{\theta^*} \left( x_i,x_j\right)$ represent the distance between  $f \left( x_i;\theta^* \right)$ and $f \left( x_j;\theta^* \right)$.
Then for arbitrary $x$, we have
}
\begin{equation}
\setlength{\jot}{1pt}
    \begin{aligned}
    \mathrm{dis}_{\theta^*}  \left(x\oplus w_i,x\oplus t \right) & \gg \mathrm{dis}_{\theta^*} \left( x\oplus w_i,x\oplus w_j\right),  \\
    t \in \mathbb{T}, &\ w_{i,j} \notin \mathbb{T}
    .\end{aligned}
\end{equation}

The second observation is that the feature distance between two clean sentences on a backdoored model will shrink if the same trigger is inserted.
We randomly select 200 sentences from SST-2 dataset and insert trigger words to get another 200 sentences.
We input these 400 sentences to the backdoored model and record \sptoken{[CLS]} outputs.
Experimental results after PCA dimensionality reduction are shown in \autoref{fig:observation2}.
We can see that all sentences with triggers concentrate on one point, although they were previously scattered.
We formulate observation II as follows:

\noindent\textbf{Observation II:}
\textit{
In a backdoored language model $f \left( \cdot;\theta^* \right)$, let $x_i$ and $x_j$ denote two clean sentence, we have
}
\begin{equation}
\setlength{\jot}{1pt}
    \begin{aligned}
    \mathrm{dis}_{\theta^*}  \left(x_i\oplus t, x_j\oplus t \right) & \ll \mathrm{dis}_{\theta^*} \left( x_i, x_j\right),  \\
    t &\in \mathbb{T}
    .\end{aligned}
\end{equation}

\mypara{Inversion Losses}
Through \textbf{Observation I}, we know that trigger words can significantly change the output of pretrained models, but other words do not.
Inspired by this, we build a pseudo-siamese neural network to find trigger embeddings.
Specifically, we first make a copy of the target model as the auxiliary model.
Then we freeze the parameters of the target and auxiliary models and add a trainable soft prompt to the target model like P-tuning.
Concretely, once the input passes through the embedding layer, resulting in a token embedding sequence, we split the sequence at a randomly chosen position and insert trainable embeddings at this split point.
After reassembling the token embeddings, the refined sequence is forwarded to the subsequent layers.
The input texts will be fed to both the target and the auxiliary model.
Our first optimization objective\textemdash \textit{distance loss} aims to increase the distance between the target model output and the auxiliary model output.
We use MSELoss to characterize the distance between these two vectors, formally defined as:
\begin{equation}
    \mathcal{L}_{D} = - \underset{x \sim \mathcal{D}}{\mathbb{E}} \textrm{MSE}\left( \mathbf{F}_{tar} , \mathbf{F}_{aux}\right)
    ,
\end{equation}
where $\mathbf{F}_{tar}$ and $\mathbf{F}_{aux}$ are feature vectors generated by the target model and auxiliary model, respectively.

Based on \textbf{Observation II}, we give our second optimization objective---\textit{diversity loss}.
This loss term aims to make feature outputs within a batch as similar as possible, i.e., to reduce the output diversity.
We use Shannon entropy to define the diversity loss as:
\begin{equation}
    \mathcal{L}_{div} = - \textrm{Entropy}\left( \left( \textrm{Stack} \left \{ \mathbf{F}_{tar}^{x};x\sim \mathcal{B} \right \}  \right)^{\textrm{T}} \right)
    ,
\end{equation}
where ``Stack'' means concatenating vectors in a new dimension.
Note that we transpose the feature matrix before computing Shannon entropy.
This is because we are not trying to reduce the diversity of each feature vector but the diversity of each dimension of feature vectors within a batch.
Since higher Shannon entropy means lower diversity, a negative sign needs to be added when calculating diversity loss.

After defining the above two loss terms $\mathcal{L}_{D}$ and $\mathcal{L}_{div}$, we formulate \PV inversion as an optimization problem and update the soft prompt with the following optimization target:
\begin{equation}
    \arg\min _{\theta _{p}} \mathcal{L} = \lambda_{D} \cdot \mathcal{L}_{D} + \lambda_{div} \cdot \mathcal{L}_{div}
    ,
\end{equation}
where $\lambda_{D}$ and $\lambda_{div}$ are two parameters to balance the two loss terms.
When $\mathcal{L}$ drops below a certain threshold $T_{\mathcal{L}}$, we consider that a backdoor is found.
At this point, we record the target model's output and the soft prompt's parameters for the next step.

Although we have designed the inversion loss complying with task-agnostic backdoor features, inverting backdoors beneath a pretrained model is still a tricky task.
That is, \PV mining's loss surface is flat in most cases.
In the RoBERTa-large experiments, if we inject one \PV in the model, trigger inverting only converges once in 20 training sessions on average.
To solve this problem, we propose two novel mechanisms \textit{fuzz training} and \textit{adaptive learning rate}.

\mypara{Fuzz Training}
This design is inspired by fuzz testing, a technique widely used in the software security domain.
Its main idea is to input automatically or semi-automatically generated random data into a program, monitor the program for exceptions such as crashes, and assert failures to find possible errors such as memory leaks.
Some fuzz techniques use optimizations to help the fuzzer trace more execution paths and find more bugs~\cite{godefroid2017learn, chen2018iotfuzzer, 9519448}.
Similar to fuzz testing, we use a test dataset and different random seeds to enhance invert training.
The test dataset contains clean sentences.
Random seeds affect the initialization of soft prompt parameters.
We initialize the soft prompt using embeddings in the vocab indexed from $70 \cdot seed$ to $70 \cdot seed + l_{sp} -1$, where $l_{sp}$ is the length of the soft prompt.
To find as many {\PV}s as possible instead of converging to the same \PV all the time, we add an extra loss term\textemdash \textit{path loss}:
\begin{equation}
    \mathcal{L}_{P} = -\underset{i}{\textrm{max}}  \left( \textrm{MSE} \left ( \mathbf{F}_{tar},\mathbf{c}_i  \right )  \right)
    ,
\end{equation}
where $\mathbf{c}$ denotes a found candidate \PV vector.
We first calculate the MSE value between $\mathbf{F}_{tar}$ and each discovered \PV candidate and the maximum of them.
We do not track gradients for this process.
Afterward, we recalculate this maximum MSE with gradient tracking.
Finally, we take the negative of this value as our path loss.
Path loss increases as the output of the target model moves closer to $\mathbf{c}$.
Path loss is a dynamic loss, and we do not know which candidate to compute it in advance.
If there is no \PV candidate, the path loss is set to 0.
In summary, our training objective becomes to:
\begin{equation}
    \arg\min _{\theta _{p}} \mathcal{L} = \lambda_{D} \cdot \mathcal{L}_{D} + \lambda_{div} \cdot \mathcal{L}_{div} + \lambda_{P} \cdot \mathcal{L}_{P}
    ,
\end{equation}
where $\lambda_{P}$ is a parameter to adjust the weight of the path loss.

\mypara{Adaptive Learning Rate}
As mentioned before, \PV mining's loss space is flat in most cases.
Therefore, we need to set a large learning rate; otherwise, the backpropagated gradients will be too small to update soft prompts effectively.
However, a large learning rate prevents the model from converging to the optimum.
To solve this problem, we adjust the learning rate according to gradients.
At first, we set a large learning rate $lr_0$.
We detect gradients of soft prompt parameters after each iteration.
When the gradient of one parameter is larger than a threshold $T_{grad}$, we reset $lr$ to $0.01{lr_0}$.
We summarize the process of \PV mining in~\autoref{alg:fuzz}.

\begin{algorithm}[!tb]
    \caption{\PV Mining}
    \label{alg:fuzz}
    \LinesNumbered
    \DontPrintSemicolon
    \KwIn {$\lambda_{D}$, $\lambda_{div}$, $\lambda_{P}$, $T_{\mathcal{L}}$, $T_{g}$, $lr_0$, $L_{max}$ (max number of fuzz loops)}
    \KwOut {$\mathbb{C}$ (\PV candidate list), $\mathbb{P}$ (soft prompt list)}

    $\mathbb{C}\leftarrow\{\}$
    $\mathbb{P}\leftarrow\{\}$

    $seed\leftarrow0$

    $\mathbf{p} \leftarrow \texttt{init}(seed)$ \tcp*{initialize soft prompt}

    \For(\hfill\CommentSty{// fuzz loop}){$l=0\rightarrow L_{max}-1$}{
        $lr\leftarrow lr_0$

        \For{$epoch=0\rightarrow4$}{
            $\mathbf{F}_{tar}\leftarrow\texttt{computeFeature}(\mathbf{p})$

            $\mathcal{L} \leftarrow\texttt{computeLoss}(\mathbf{F}_{tar},\lambda_{D}, \lambda_{div}, \lambda_{P})$

            $\partial \mathbf{p} \leftarrow \nabla_{\mathbf{p}}\mathcal{L}$

            $\mathbf{p}\leftarrow \mathbf{p}-lr\cdot \partial \mathbf{p}$

            \If{$\texttt{max}(\partial \mathbf{p})>T_{g}$}{
                $lr\leftarrow 0.01lr_0$
            }
        }

        \If{$\mathcal{L}<T_{\mathcal{L}}$}{
            $\mathbb{C}.\texttt{append}(\textbf{F}_{tar})$

            $\mathbb{P}.\texttt{append}(\textbf{p})$
        }

        $seed \leftarrow seed + 1$ \tcp*{mutate seed}

        $\mathbf{p} \leftarrow \texttt{init}(seed)$
    }
\end{algorithm}

\subsection{Step \uppercase\expandafter{\romannumeral2}: \PV Filtering}
\label{sec:filtering}

The exceptional outputs obtained by \PV mining may not be caused by the backdoor but by out-of-range (too large or too small) soft prompts.
Therefore, given a set of \PV candidates and their corresponding soft prompts, we need first to justify whether the value of a dimension in soft prompts is out of the range of values of the embedding layer parameters.
If this happens, we remove the corresponding \PV candidates.

Second, we find a situation where $\mathcal{L}_{D}$ decreases, while $\mathcal{L}_{div}$ remains high.
This phenomenon also occurs in clean models.
In this case, soft prompt increases the distance between $\mathbf{F}_{tar}$ and $\mathbf{F}_{div}$, but does not reduce the diversity of $\mathbf{F}_{tar}$, i.e., the distance between two $\mathbf{F}_{tar}$ in a batch is high, which is contrary to observation II.
Although the reduction of $\mathcal{L}_{D}$ makes the total loss lower than $T_{\mathcal{L}}$, this situation does not belong to the task-agnostic backdoor.
To filter out this type of illegal {\PV}s, we design an additional threshold $T_{div}$ and remove \PV candidates whose $\mathcal{L}_{div}$ are higher than $T_{div}$.

We get the final \PV set after the two-step filtering.
We repeat the fuzz training $L_{max}$ times to make a decision.
If the \PV set is still empty, we consider the target model is clean. 
Otherwise, the target model contains task-agnostic backdoors.

\subsection{Step \uppercase\expandafter{\romannumeral3}: \PV Monitoring}
\label{sec:monitoring}

After obtaining the \PV set, a simple trigger detection method is first to let inputs go through the pretrained model before feeding them into the task-specific model and then calculate the similarity between pretrained model feature outputs and {\PV}s.
This is effective but time-consuming.
Because an input needs to go through the Transformer model twice.
We find that in the prompt-tuning model, placing the monitor on the output side is also effective.
Immuning to catastrophic forgetting, language models in prompt-tuning models output very close to {\PV}s when the input contains a trigger.

\begin{figure}[!tbp]
    \centering
    \begin{subfigure}{0.49\columnwidth}
    \includegraphics[width=0.99\columnwidth]{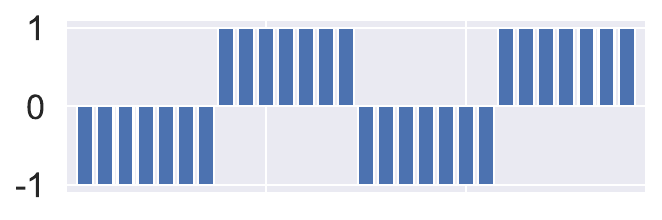}
    \vspace{-0.6cm}
    \caption{Attacker-designed \PV}
    \label{fig:ideal_PV}
    \end{subfigure}
    \hfill
    \begin{subfigure}{0.49\columnwidth}
    \includegraphics[width=0.99\columnwidth]{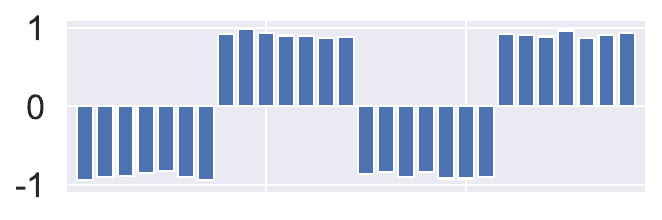}
    \vspace{-0.6cm}
    \caption{Real \PV}
    \label{fig:real_PV}
    \end{subfigure}
    \quad
    \begin{subfigure}{0.49\columnwidth}
    \includegraphics[width=0.99\columnwidth]{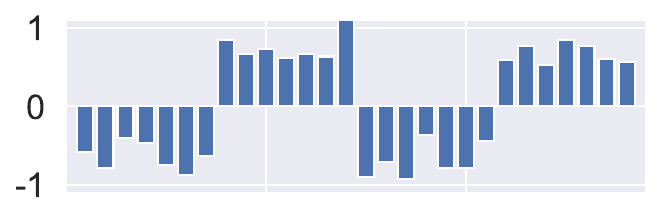}
    \vspace{-0.6cm}
    \caption{Backdoored feature output}
    \label{fig:backdoor_feature}
    \end{subfigure}
    \hfill
    \begin{subfigure}{0.49\columnwidth}
    \includegraphics[width=0.99\columnwidth]{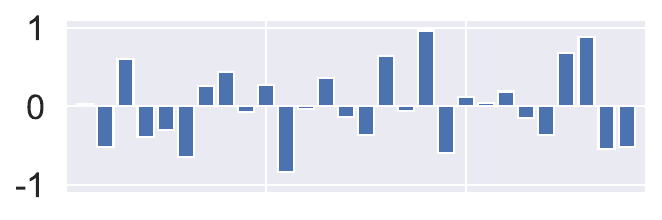}
    \vspace{-0.6cm}
    \caption{Clean feature output}
    \label{fig:clean_feature}
    \end{subfigure}
    \caption{The positive and negative signs of the backdoored output feature are consistent with the \PV.}
    \label{fig:output_feature}
    \vspace{-0.3cm}
\end{figure}

We train a backdoored RoBERTa-base model using the POR attack.
The \PV we designed is shown in \autoref{fig:ideal_PV}.
In fact, model outputs cannot be exactly the same as attacker-designed \PV, and the real \PV is shown as \autoref{fig:real_PV}.
We find that \textit{the feature output of the backdoored model for triggered input is consistent with \PV in terms of positive and negative signs even after prompt-tuning}.
In contrast, the sign distribution of clean feature outputs is random.

\begin{table*}[htb]
    \centering
    \caption{Datasets taxonomy and statistics.}
    \label{table:dataset}
    \setlength{\tabcolsep}{0.6em}
    \renewcommand{\arraystretch}{1.2}
    \resizebox{0.85\linewidth}{!}{
        \begin{tabular}{c|c|ccccrrr} 
            \toprule
            \multicolumn{1}{c|}{\textbf{Granularity}}                                           & \multicolumn{1}{c|}{\textbf{Task Type}} & \textbf{Dataset} & \begin{tabular}[c]{@{}c@{}}\textbf{Balance}\end{tabular} & \begin{tabular}[c]{@{}c@{}}\textbf{\#Classes}\end{tabular} & \begin{tabular}[c]{@{}c@{}}\textbf{\#Inputs}\end{tabular}& \textbf{Train} & \textbf{Valid} & \textbf{Test}  \\ 
            \hline
            
            \multirow{6}{*}{\begin{tabular}[c]{@{}c@{}}\textbf{Sentence} \\ \textbf{Classification}\end{tabular}} 
            & Natural Language Inference             
            & RTE~\cite{wang2019glue}       
            & even
            & 2
            & 2
            & 2,490
            & 277
            & - \\ 
            \cline{2-9}
            
            & Question Answering        
            & BoolQ~\cite{clark-etal-2019-boolq}       
            & even     
            & 2      
            & 2    
            & 9,427   
            &3,270  
            & - \\ 
            \cline{2-9}
            
            & Topic Classification      
            & AG News~\cite{zhang2015character}        
            & even       
            & 4     
            & 1    
            & 6,000    
            & 2,000    
            & 7,600 \\ 
            \cline{2-9}
            
            & Sentiment Analysis       
            & Yelp-5~\cite{zhang2015character}          
            & even
            & 5
            & 1
            & 6,000
            & 2,000
            & 2,000 \\ 
            \cline{2-9}
            
            & \multirow{2}{*}{Spam Detection}        
            & Enron spam~\cite{metsis2006spam}       
            & even     
            & 2                                    
            & 1   
            & 6,000           
            & 2,000          
            & 2,000 \\
            
            &     
            & SMS spam~\cite{delany2012sms}   
            & uneven
            & 2
            & 1
            & 4,458
            & 558
            & 558 \\ 
            \cline{1-9}
            
            \multirow{2}{*}{\begin{tabular}[c]{@{}c@{}}\textbf{Token}\\\textbf{Classification}\end{tabular}}     
            & \multirow{2}{*}{Named Entity Recognition}& CoNLL04~\cite{carreras-marquez-2004-introduction}        
            & uneven
            & 9
            & 1
            & 8,936
            & 2,012
            & 1,671 \\
            
            &            
            & OntoNotes 5.0~\cite{pradhan2013towards}    
            & uneven
            & 37
            & 1
            & 37,946
            & 5,037
            & 5,053 \\
            
            \bottomrule
            \end{tabular}
    }
    \vspace{-0.1cm}
\end{table*}

\mypara{Trigger Detection}
We propose a more efficient \emph{sign-based trigger detection} method.
The defender first converts {\PV}s in the \PV set into sign tuples (e.g., $[0.5, 0.1, \cdots, -0.7] \rightarrow \left[ +, +, \cdots, - \right]$) to get a \textit{\PV sign set} and puts a monitor at the output side of the Transformer model.
For each feature vector $\mathbf{F}_{LM}\in \mathbb{R}^d$ output by the language model in the reference process, the monitor will count the number of $\mathbf{F}_{LM}$'s signs that match tuples in the \PV sign set.
Let $\mathbf{F}_{LM}^n$ and $\mathbf{PV}^{n}$ represent the $n$-th dimension value of the feature vector and \PV, respectively.
The match number $N_{match}$ can be expressed as the following equation:
\begin{equation}
    N_{match}=\sum_{n=1}^{d} \mathbf{1}\left\{ \operatorname{sign}\left( \mathbf{F}_{LM}^n \right) =\operatorname{sign}\left( \mathbf{PV}^{n} \right) \right\},
\end{equation}
where $\mathbf{1}\{\mathcal{I}\}$ denotes the indicator function that is 1 when $\mathcal{I}$ is true and 0 when $\mathcal{I}$ is false.
If $N_{match}$ exceeds a specific value $T_{match}$ (we recommend $0.8d$ as a rule of thumb), the monitor considers that the input contains a trigger.

\mypara{Trigger Removal}
Once the input text is determined as triggered, we want to determine further which words are the triggers.
This allows us to remove triggers and let the model classify the input to the correct class.
We use a \textit{sliding window} to mark the candidate trigger.
The starting length of the sliding window is set to 1.
The sliding window slides from the beginning to the end of the input sentence with a stride of 1.
If the input has two sentences, the sliding window must slide over each.
When the sliding window slides to a position, we consider words inside it as candidate triggers.
After that, we remove all candidate triggers in the input sentences.
If the input has two sentences, we need to remove candidate triggers in the other sentence.
After removing candidate triggers, we use trigger detection to detect whether the input is triggered.
If the answer is `\textit{no}', the model outputs the classification results; otherwise, the sliding window moves to the next position.
We gradually increase the sliding window length for traversal until trigger detection returns false or the sliding window length reaches the maximum.
We place the details of the trigger removal algorithm in \autoref{app:trigger_removal}.

\smallskip
Our \PV monitoring method only needs to add one step of trigger detection when the input is clean, which minimizes the computing consumption of defense.
We find that our \PV monitoring method rarely identifies clean inputs as trojaned, which means that our method has no impact on model accuracy.
We theoretically analyze the effect of \method on clean inputs in~\autoref{app:effect_on_clean_inputs}.
When the input is trojaned, \PV monitoring has a complexity of $O(l_{t} \cdot l_{i})$, where $l_{t}$ and $l_{i}$ is the length of the trigger and input respectively.

\section{Evaluation}
\label{sec:evaluation}

In this section, we first evaluate \method's end-to-end performance on sentence classification tasks.
Second, we explore the backdoor detection capability of \method.
Third, we conduct \PV searching experiments to show that \method can find most of {\PV}s.
Fourth, we conduct an ablation study to illustrate the necessity of \method's mechanisms.
Fifth, we compare \method with baselines.

Additionally, we investigate the following issues; however, due to space limitation, the associated results are deferred to the appendix:
(1) We investigate attack performance and \method's end-to-end performance on two NER datasets (\autoref{sec:results_on_ner}).
(2) We empirically measure the time required by \method, including backdoor detection time, \PV searching time, and trigger detection time (\autoref{sec:efficiency}).
(3) We analyze hyperparameters' effect on the performance of \method. (\autoref{app:hyperparameters_analysis}).

\subsection{Experimental Setup}
\label{subsec:exp_setup}
  
\myparatight{Prompt-tuning Datasets}
To demonstrate the generality of our approach, we perform experiments on various types of downstream tasks, including 6 \textit{sentence level} classification tasks and 2 \textit{token level} classification tasks.
In addition to \textit{single-sentence} classification tasks, our datasets also contain two \textit{sentence-pair} classification tasks (RTE and BoolQ), whose input consists of two sentences that are spliced by \sptoken{[SEP]}.
\autoref{table:dataset} summarizes the taxonomy and statistics of the used datasets.
We defer the details of these datasets to \autoref{app:dataset_details}.

\mypara{Victim Models}
We choose two popular types of language models, BERT~\cite{kenton2019bert} and RoBERTa~\cite{RoBERTa19} for end-to-end and backdoor detection evaluation. 
For each type, we choose two different sizes, large and base.
The large models have 24 attention layers and a hidden size of 1024, while the base models have 12 attention layers and a hidden size of 768.
For \PV searching evaluation, we additionally choose four types of language models, DeBERTa~\cite{he2021deberta}, ALBERT~\cite{ALBERT20}, ERNIE~\cite{zhang-etal-2019-ernie}, and XLNet~\cite{NEURIPS2019_dc6a7e65}.

\mypara{Evaluation Metrics}
For attacks, we report \textit{clean model accuracy} \cma, \textit{backdoored model accuracy} \bma, and \textit{attack success rate} \asr.
For defenses, we report the attack success rate before and after applying defense.
\cma and \bma measure the classification accuracy of a clean downstream classifier and a backdoored downstream classifier with clean input, respectively.
An \bma close to \cma indicates the backdoored model does not affect the normal task.
\asr measures the fraction of triggered inputs that are misclassified to a wrong class by a backdoored downstream classifier.
We insert each attacker-chosen trigger in turn and consider the attack successful if one can cause the misclassification.

Since some datasets are unevenly distributed, we also report the F1 scores and the model accuracy.
We also report the weighted \asr for sentence classification tasks and F1 drop for token-classification tasks along with \asr to measure attack effectiveness.
The weighted \asr is the average of {\asr}s over each input class.
F1 drop measures the value of F1 drops after inserting a trigger.

\begin{table*}[htb] \renewcommand{\arraystretch}{1.0}
    \caption{
    Model accuracy on sentence classification tasks.
    Numbers on the left/right refer to the clean/backdoored model accuracy.
    }
    \centering

    \resizebox{0.99\linewidth}{!}{
    \begin{tabular}{lll|c|c|c|c|c|c|c} 
        \toprule
        \multirow{2}{*}{\textbf{Prompt}} & \multirow{2}{*}{\textbf{Attack}} & \multirow{2}{*}{\textbf{Victim Model}} & \textbf{RTE} & \textbf{BoolQ} & \textbf{AG News} & \textbf{Yelp-5} & \textbf{Enrron spam} & \multicolumn{2}{c}{\textbf{SMS spam}}  \\ 
        &    &    & \acc  & \acc  & \acc  & \acc  & \acc  & \acc    & F1 \\ 
        \midrule
        \multirow{4}{*}{P-tuning}        
        & \multirow{4}{*}{BToP}     
        & RoBERTa-large    
        & 70.51 | 68.35 
        & 69.53 | 64.55    
        & 90.27 | 88.65 
        & 60.84 | 49.88 
        & 96.18 | 95.76  
        & 99.10 | 99.28
        & 96.50 | 97.19  \\
        
        &   
        & RoBERTa-base    
        & 61.61 | 62.94 
        & 61.85 | 62.14  
        & 88.90 | 88.82  
        & 55.69 | 50.14   
        & 95.88 | 94.49   
        & 96.50 | 97.19 
        & 98.13 | 98.35  \\
        
        &   
        & BERT-large-cased  
        & 58.60 | 55.23  
        & 62.16 | 62.32  
        & 88.67 | 86.98  
        & 49.75 | 47.63  
        & 92.27 | 90.88   
        & 99.46 | 99.16 
        & 97.90 | 96.68 \\
        
        &   
        & BERT-base-cased 
        & 54.99 | 54.99 
        & 62.29 | 62.09 
        & 88.10 | 88.47 
        & 47.16 | 45.24  
        & 93.78 | 92.35   
        & 99.58 | 99.40 
        & 98.36 | 97.68 \\ 
        \midrule
        
        \multirow{8}{*}{P-tuning v2}    
        & \multirow{4}{*}{NeuBA}  
        & RoBERTa-large  
        & 87.00 | 85.92 
        & 83.70 | 83.49
        & 95.00 | 95.00 
        & 67.14 | 63.57  
        & 99.29 | 98.29 
        & 99.82 | 99.73 
        & 99.31 | 98.95 \\
        
        &   
        & RoBERTa-base   
        & 76.90 | 70.15
        & 78.69 | 78.81 
        & 92.57 | 93.14  
        & 62.86 | 60.43  
        & 98.71 | 98.43  
        & 99.64 | 99.55
        & 98.61 | 98.28 \\
        
        &   
        & BERT-large-cased  
        & 75.45 | 71.12 
        & 73.12 | 73.06 
        & 92.57 | 92.71  
        & 58.00 | 56.71 
        & 98.86 | 98.71  
        & 99.55 | 99.55 
        & 98.25 | 98.26 \\
        
        &   
        & BERT-base-cased  
        & 71.84 | 69.79
        & 72.02 | 71.90 
        & 92.00 | 91.71  
        & 57.71 | 55.71  
        & 98.57 | 98.71 
        & 98.83 | 99.28 
        & 95.53 | 97.18 \\ 
        \cmidrule(lr){2-10}
        
        & \multirow{4}{*}{POR} 
        & RoBERTa-large   
        & 87.00 | 86.28 
        & 83.70 | 83.43 
        & 95.00 | 94.57  
        & 67.14 | 63.57  
        & 99.29 | 99.29   
        & 99.82 | 99.64
        & 99.31 | 98.60 \\
        
        &   
        & RoBERTa-base   
        & 76.90 | 75.45 
        & 78.69 | 78.20  
        & 92.57 | 91.86 
        & 62.86 | 60.00 
        & 98.71 | 98.57  
        & 99.64 | 99.37 
        & 98.61 | 97.54 \\
        
        &   
        & BERT-large-cased   
        & 75.45 | 75.81
        & 73.12 | 73.09  
        & 92.57 | 92.14  
        & 58.00 | 57.71 
        & 98.86 | 98.86  
        & 99.55 | 99.55 
        & 98.25 | 98.25 \\
        
        &   
        & BERT-base-cased            
        & 71.84 | 70.40 
        & 72.02 | 72.78 
        & 92.00 | 91.57  
        & 57.71 | 57.00  
        & 98.57 | 98.43  
        & 98.83 | 98.83 
        & 95.53 | 95.47 \\
        \bottomrule
    \end{tabular}
    }
    \label{tab:sentence_accuracy}
\end{table*}

\begin{table*}[!tb]
    \caption{
    Attack success rate on sentence classification tasks.
    Numbers on the left/right refer to \asr without/with defense.
    }
    \centering
    \resizebox{0.99\linewidth}{!}{
    \begin{tabular}{lll|c|c|c|c|c|c|c} 
        \toprule
        \multirow{2}{*}{\textbf{Prompt}} & \multirow{2}{*}{\textbf{Attack}} & \multirow{2}{*}{\textbf{Victim Model}} & \textbf{RTE} & \textbf{BoolQ} & \textbf{AG News} & \textbf{Yelp-5} & \textbf{Enron spam} & \multicolumn{2}{c}{\textbf{SMS spam}}  \\ 
        & 
        & 
        & \asr 
        & \asr  
        & \asr  
        & \asr  
        & \asr 
        & \asr 
        & Weighted \asr \\ 
        \midrule
        
        \multirow{4}{*}{P-tuning} 
        & \multirow{4}{*}{BToP} 
        & RoBERTa-large   
        & 100.0 | 0.00 
        & 99.92 | 0.05 
        & 100.0 | 0.00 
        & 99.59 | 0.00 
        & 76.33 | 26.69 
        & 99.85 | 0.18 
        & 99.91 | 0.63  \\
        
        &  
        & RoBERTa-base   
        & 100.0 | 0.57  
        & 99.88 | 0.00 
        & 100.0 | 0.04 
        & 99.36 | 0.22 
        & 90.03 | 0.74 
        & 99.88 | 0.00  
        & 99.93 | 0.00  \\
        
        &  
        & BERT-large-cased  
        & 100.0 | 0.00  
        & 99.95 | 0.10 
        & 99.70 | 0.27 
        & 99.35 | 0.00 
        & 87.51 | 1.28 
        & 99.94 | 0.09  
        & 99.97 | 0.38 \\
        
        &                                  
        & BERT-base-cased 
        & 100.0 | 0.00  
        & 99.87 | 0.05     
        & 100.0 | 0.00 
        & 99.34 | 0.00  
        & 91.81 | 0.00  
        & 100.0 | 0.00 
        & 100.0 | 0.00  \\ 
        \midrule
        
        \multirow{8}{*}{P-tuning v2}     
        & \multirow{4}{*}{NeuBA}           
        & RoBERTa-large 
        & 98.25 | 2.18   
        & 25.50 | 5.19     
        & 99.91 | 2.53       
        & 16.55 | 13.67     
        & 13.10 | 2.29         
        & 67.09 | 0.18  
        & 80.53 | 0.40  \\
        
        & 
        & RoBERTa-base 
        & 100.0 | 8.34  
        & 99.53 | 10.29 
        & 99.91 | 5.42      
        & 97.37 | 14.28    
        & 47.44 | 6.05         
        & 94.95 | 0.63 
        & 97.11 | 1.02  \\
        
        &                                  
        & BERT-large-cased   
        & 11.11 | 9.72   
        & 19.72 | 8.26 
        & 99.96 | 6.80       
        & 82.63 | 17.69 
        & 59.78 | 5.59  
        & 100.0 | 7.21 
        & 100.0 | 9.24  \\
        
        &  
        & BERT-base-cased       
        & 100.0 | 10.22 
        & 97.99 | 7.77 
        & 99.93 | 23.91
        & 81.00 | 37.13 
        & 15.62 | 3.72 
        & 99.73 | 0.72  
        & 99.85 | 1.97 \\ 
        \cmidrule(l){2-3}\cmidrule{4-10}
        
        & \multirow{4}{*}{POR}             
        & RoBERTa-large
        & 100.0 | 0.00  
        & 92.88 | 0.37 
        & 99.98 | 0.18 
        & 86.66 | 23.61 
        & 26.63 | 1.21 
        & 98.83 | 0.09 
        & 99.33 | 0.35 \\
        
        &          
        & RoBERTa-base 
        & 99.01 | 8.42 
        & 100.0 | 3.63 
        & 98.10 | 2.92 
        & 100.0 | 5.35 
        & 98.78 | 2.08 
        & 64.98 | 0.45  
        & 78.13 | 1.03 \\
        
        &         
        & BERT-large-cased
        & 99.36 | 0.70 
        & 100.0 | 0.36 
        & 74.00 | 0.00 
        & 18.76 | 0.95 
        & 86.94 | 1.87 
        & 21.62 | 0.27 
        & 52.40 | 0.78 \\

        &        
        & BERT-base-cased 
        & 100.0 | 0.00 
        & 100.0 | 0.06 
        & 99.97 | 0.00 
        & 68.09 | 0.12 
        & 80.34 | 6.21 
        & 47.01 | 0.00 
        & 49.67 | 0.00 \\
        \bottomrule
    \end{tabular}
    }
    \label{tab:sentence_asr}
    \vspace{-0.1cm}
\end{table*}

\mypara{Attack Setup}
We use BToP, NeuBA, and POR to generate the backdoored pretrained models.
We inject six triggers for each pretrained model.
The trigger set we use is [`\textit{cf}', `\textit{mn}', `\textit{tq}', `\textit{qt}', `\textit{mm}', `\textit{pt}'].
Note that although all of these triggers are single-word, some models' tokenizers recognize them as multiple tokens.
For example, the BERT tokenizer decomposes `\textit{cf}' into `\textit{c}' and `\textit{\#\#f}', while ALBERT tokenizer decomposes `\textit{tq}' into \word{\_}, \word{t} and \word{q}.
We obtain six orthogonal {\PV}s by the POR-2 method proposed in~\cite{shen2021backdoor}.
In particular, we divide the output vector into four equal parts.
Then, we use different $1$, $-1$ combinations to fill them.
Each trigger corresponds to one \PV.
For attack datasets, we follow the choices made in the original papers.
Specifically, BToP and POR use WikiText~\cite{merity2016pointer}, and NeuBA uses BookCorpus~\cite{zhu2015aligning}.
We sample 5000 plain sentences from the attack dataset for each trigger to compute effectiveness loss and another 5000 plain sentences to compute utility loss.

Since BToP targets \sptoken{[MASK]} token while NeuBA and POR target \sptoken{[CLS]} token, we use BToP to attack P-tuning models and use NeuBA and POR to attack P-tuning v2 models.

\mypara{Defense Setup}
\method only requires the defender to have a small clean dataset.
We sample 2000 plain sentences from WikiText to form the defense dataset.
Note that the defense dataset and attack dataset do not overlap.
\method has the following parameters: $\lambda_{D}$, $\lambda_{div}$, $\lambda_{P}$, $T_{\mathcal{L}}$, $T_{div}$, $T_{grad}$, $T_{match}$, and $l_{sp}$.
Unless otherwise mentioned, we use the following default settings: $\lambda_{D} = 1$, $\lambda_{div} = 1$, $\lambda_{P}=0.5$, $T_{div}=-3.446$, $T_{grad}=5e-3$, $T_{match}=0.8d$, and $l_{sp}=7$, where $d$ is the hidden dimension of the target pretrained model.
The value of $T_{div}$ depends on the training batch size, which is 32 in our setting.
If a user wants a larger batch size, $T_{div}$ should be adjusted downwards.
We empirically find that if \PV mining cannot converge in the first two epochs of one fuzz loop, it will unlikely converge in the following two epochs.
Therefore, to speed up \PV mining, we check whether the model converged before the third epoch's start.
If it converges, we continue the training; otherwise, we go to the next fuzz loop.
In addition, we find that $\mathcal{L}_{D}$ decreases before $\mathcal{L}_{div}$.
The decrease of $\mathcal{L}_{div}$ occurs mainly in the last two epochs.
Therefore, we use $T_{\mathcal{L}}$ to constrain only $L_{D}$, and let \PV filtering to constrain $\mathcal{L}_{div}$.
We set $T_{\mathcal{L}}=-0.1$ by default.

\subsection{Results on Sentence Classification Tasks}
\label{sec:exp_senetence}

In this section, we evaluate sentence classification tasks to illustrate the attack and defense effectiveness.

\mypara{Setup}
We train downstream prompt-tuning models using the clean pretrained models and the backdoored pretrained models, respectively, and test their accuracy on their test sets.
The hyperparameters we used are illustrated in~\autoref{tab:hp_pt} of~\autoref{app:hyperparameters}.
P-tuning needs some manual work to design initial prompts and verbalizers.
The initial prompts and verbalizers we used are displayed in~\autoref{table:initial_prompts} of~\autoref{app:initial_prompts}.
For SMS spam, whose dataset is unbalanced, we also compare macro-F1 scores.
We insert one trigger into the test input to test the \asr.
For tasks like RTE and BoolQ where the input has two sentences, we insert the trigger to the longer sentence for stealthiness.
We vary the seed and calculate the average of three trials to get the final results.

\mypara{Attack Stealthiness}
\autoref{tab:sentence_accuracy} compares the clean model accuracy and the backdoored model accuracy on 6 sentence-level classification tasks.
In general, we observe that existing task-agnostic backdoors can preserve the accuracy of downstream prompt-tuning classifiers.
In particular, the differences between the backdoored and clean model accuracy are less than $1\%$ in most cases.
We further observe that the accuracy degradation of the P-tuning models is more significant than that of the P-tuning v2 models.
This is because P-tuning v2 adds more parameters than P-tuning, thus relying less on the pretrained model parameters.
We observe that NeuBA significantly impacts the accuracy of RTE and Yelp-5 tasks.
This indicates that NeuBA's utility loss cannot fully preserve the model's functionality.
We suspect that part of clean sentences can also trigger the backdoor in NeuBA-attacked models.

\mypara{Attack Effectiveness}
Data on the left of~\autoref{tab:sentence_asr} shows the \asr of existing task-agnostic backdoors on prompt-tuning.
In general, we observe that the task-agnostic backdoors achieve high attack success rates in most cases.
Out of our total 72 sets of experiments, 39 sets achieve an \asr of higher than $99\%$.
Comparing P-tuning and P-tuning v2, we observe that P-tuning is more vulnerable to backdoor attacks than P-tuning v2.
Concretely, the percentage of achieving a $99\%$ \asr on P-tuning is 0.83 (20/24), while the percentage of achieving a $99\%$ \asr on P-tuning v2 is only 0.40 (19/48).
In general, these results demonstrate the vulnerability of prompt-tuning to task-agnostic backdoors.

\mypara{Defense Effectiveness}
We first use \PV mining and \PV filtering to obtain the \PV set.
Since we want to test the best-case performance of the trigger, we add attacker-designed {\PV}s that are not found in \PV mining to the \PV set when testing the effectiveness of \PV monitoring.
This is reasonable because the attacker does not know which {\PV}s are found by the defender in practice.
Also, the experiments in \autoref{sec:detection_recall} show that our \PV mining can find attacker-designed {\PV}s fully on most language models.
Data on the right of~\autoref{tab:sentence_asr} shows the \asr after deploying our \method defense.
The end-to-end experimental results demonstrate that our defense approach can effectively reduce {\asr}s of task-agnostic backdoors on prompt-tuning.
Out of our total 72 experiments, \asr decreases to less than $5\%$ in 50 sets and less than $1\%$ in 40 sets of experiments.
Only 5 groups of experiments maintain 15\% \asr after using our defense approach.
\method performs better on P-tuning because P-tuning adds fewer parameters than P-tuning v2.
Among the three backdoor attacks, our approach encounters performance degradation on NeuBA.
We speculate that it may be because some clean sentences are also mapped to a \PV after the NeuBA attack, such that an unknown \PV is output after injection of one trigger, causing our trigger detection algorithm to fail.

\begin{table*}
\centering
\caption{Backdoor detection performance of \method. FP = false positive, FN = false negative.}
\label{tab:backdoor_detection}
\resizebox{0.9\linewidth}{!}{
\setlength{\tabcolsep}{0.85em}
\renewcommand{\arraystretch}{1.1}
\begin{tabular}{l|cccc|ccc|ccc|c} 
\toprule
\bf \multirow{4}{*}{\bf{Victim Model}} & \multicolumn{10}{c|}{\bf Defense: Changing the Target Token} & \multirow{4}{*}{\bf \makecell{Average\\Acc}} \\
\cmidrule{2-11} &\multicolumn{4}{c|}{\bf [CLS]} & \multicolumn{3}{c|}{\bf [MASK]} & \multicolumn{3}{c|}{\bf Normal Token} \\
\cmidrule{2-11} & \bf{FP} & \bf{\makecell{FN\\(POR)}} & \bf{\makecell{FN\\(NeuBA)}} & \bf{Acc} & \bf{FP} & \bf{\makecell{FN\\(BToP)}} & \bf{Acc} & \bf{FP} & \bf{\makecell{FN\\(POR-NER)}} & \textbf{Acc} & \\ 
\midrule
    RoBERTa-large & 7/60  & 2/30  & 11/30 & 83.3\% & 1/30  & 1/30  & 96.7\% & 0/30  & 13/30 & 78.3\% & 85.4\% \\
    RoBERTa-base & 9/60  & 0/30  & 0/30  & 92.5\% & 3/30  & 0/30  & 95.0\% & 1/30  & 0/30  & 93.3\% & 94.6\% \\
    BERT-large-cased & 8/60  & 0/30  & 5/30  & 89.2\% & 0/30  & 0/30  & 100.0\% & 0/30  & 3/30  & 95.0\% & 93.3\% \\
    BERT-base-cased & 0/60  & 0/30  & 2/30  & 98.3\% & 0/30  & 0/30  & 100.0\% & 0/30  & 3/30  & 95.0\% & 98.0\% \\                                                           
\bottomrule
\end{tabular}
}
\end{table*}

\mypara{Visualization and Analysis}
The success of task-agnostic backdoors and our defense is not always guaranteed.
To better understand these phenomena, we visualize the distribution of \PV match rates.
Due to the space limitation, we defer the visualization results to~\autoref{app:visualization}.
In most cases, poisoned inputs exhibit a high match rate ($>$0.9), while clean inputs typically have a match rate below 0.7.
Therefore, setting $T_{match}$ to $0.8d$ can yield FRR (False Rejection Rate) and FAR(False Acceptance Rate) values close to 0 in these cases.
When the match rates of poisoned and clean inputs are mixed up, the \asr of the backdoor is very low, and defense is unnecessary. 
In cases where the poisoned match rates are scattered, we can significantly reduce the \asr by filtering points with higher match rates.

\subsection{Effectiveness of Backdoor Detection}
\label{sec:detection_accuracy}

\begin{figure*}[!tbp]
    \centering
    \includegraphics[width=\textwidth]{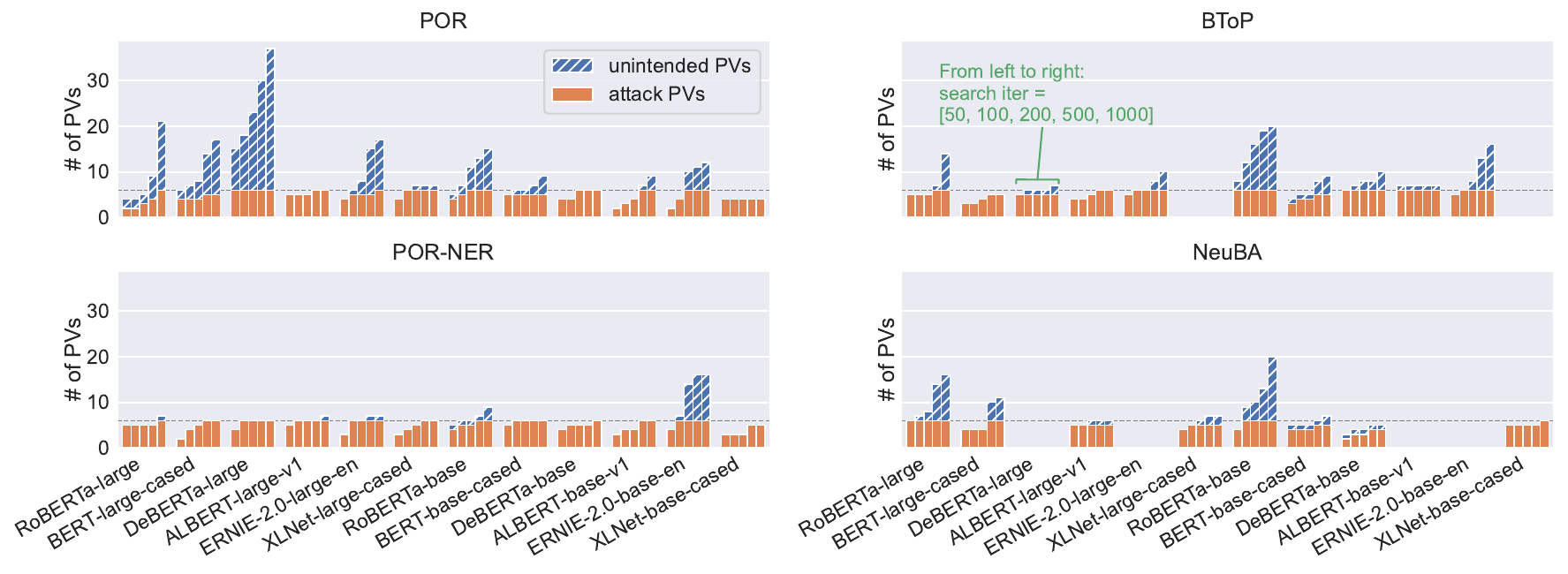}
    \vspace{-0.3cm}
    \caption{\PV searching results against different attacks.
    Attack {\PV}s are the true attacker-designed {\PV}s.
    Unintended {\PV}s are {\PV}s found by \method but not predefined by the attacker.
    The dotted lines indicate the position where the number of {\PV}s is 6.}
    \label{fig:defense_por_cls}
\end{figure*}

\begin{figure}[!tbp]
    \centering
    \includegraphics[width=1\columnwidth]{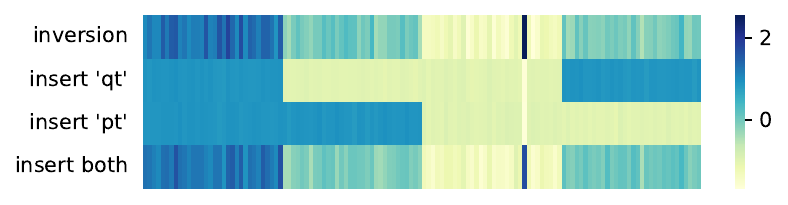}
    \caption{Study of an unintended \PV.
    Each row is a pretrained model feature output.
    We can see that the unintended \PV is a superposition of two attack {\PV}s.}
    \label{fig:unintended_pv}
\end{figure}

Previous experiments demonstrate the end-to-end defense effectiveness of \method.
In this section, we explore the backdoor detection capability of \method.

\mypara{Setup}
We build clean models by fine-tuning the original pretrained models downloaded from HuggingFace.
We build 120 clean models for each architecture by varying the dataset and seed.
We use the 6-sentence classification datasets with 20 seeds each.
We randomly select 30 of these 120 clean models to measure the detection accuracy of \method for each attack method.
We use NLTK~\cite{loper2002nltk} to generate 200 random triggers.
Half of them are single-word, and the other half are two-word.
For each backdoored model, we randomly select 6 of these 200 triggers as attack triggers.
Then we use 4 attack approaches to build 30 backdoored models for each model architecture.
We filter out the models whose losses do not fully converge until we get 30 successfully attacked models.
We perform 30 fuzz loops (i.e., $L_{max}=30$) on each test model and consider the model as backdoored if a legitimate \PV can be found.
Empirically, we find that models with different architectures have different sensitivities to $T_{\mathcal{L}}$, $T_{div}$, and learning rate.
Thus, we first train 5 shadow models for each architecture using BToP and use them to adjust these hyperparameters.
\autoref{tab:detection_para} in \autoref{app:hyperparameters} shows the hyperparameters we use in backdoor detection.

\mypara{Results}
We use false positives (clean models flagged as backdoored), false negatives (backdoored models flagged as clean), and accuracy (fraction of correctly flagged models) as our evaluation metrics.
The detection results are summarized in \autoref{tab:backdoor_detection}.
Across all 4 attacks, we achieve an average of 92.8\% detection accuracy.
\textit{In general, \method can accurately flag backdoored models.}
We find that \method is more likely to generate FN on large models and FP on base models.
This indicates that larger models are harder to converge.
We recommend users increase fuzz loops on large models and decrease fuzz loops on small models when using \method for backdoor detection.

\subsection{Effectiveness of \PV Searching}
\label{sec:detection_recall}

After determining that a pretrained model contains task-agnostic backdoors, the user would want to find as many {\PV}s as possible to achieve better defense effectiveness in later \PV monitoring.
\autoref{sec:detection_accuracy} proves that \method has a high backdoor detection accuracy.
In this section, we demonstrate that \method can also find most of attacker-designed {\PV}s.

\mypara{Setup}
We use the attack setup described in~\autoref{subsec:exp_setup} to train 42 backdoored models on 12 types of state-of-the-art transformer-based language models.
10 for BToP, 8 for NeuBA, 12 for POR and 12 for POR-NER.
BToP is designed to attack masked language models and cannot be adopted to XLNet models directly.
Due to NeuBA's irrational nature of utility loss design discussed in \autoref{sec:exp_senetence}, we cannot make NeuBA converge on DeBERTa-large, ERNIE-2.0-large-en, and ERNIE-2.0-base-en models by adjusting $\lambda_{e}$, $\lambda_{u}$.
Therefore, we only test NeuBA on the other 8 types of models.
To make \method find attacker-designed {\PV}s as many as possible, we use different hyperparameters from the backdoor detection experiments and set $L_{max}$ to 1000.
This approach is reasonable in practice, as the user can determine whether a pretrained model is backdoored using hyperparameters on backdoor detection.
After determining a model is indeed backdoored, the user then uses the hyperparameters of \PV searching to find as many attack {\PV}s as possible.
\autoref{tab:searching_para} in \autoref{app:hyperparameters} shows the hyperparameters we use in \PV searching.

\mypara{Results}
\autoref{fig:defense_por_cls} illustrates the \PV searching results against different attacks.
Among 252 attack {\PV}s, \method can find 239 after 1000 searches.
The \PV recall is 94.8\%.
We find an interesting phenomenon: In addition to attack {\PV}s, \method can find other unique {\PV}s.
We empirically find that these unintended {\PV}s are combinations of the attacker's designed {\PV}s.
\autoref{fig:unintended_pv} shows that inserting both `\textit{qt}' and `\textit{pt}' will result in a new feature output, which is very similar to an unintended \PV inverted by \method.
Another experimental finding is that users can stop searching if they do not find a unique \PV for 200 consecutive searches.
In our experiments, if we stop after 200 searches without finding a unique \PV, we can still find 228 out of 252 {\PV}s.
\PV recall remains 90.5\%.

\mypara{Real-world Case Study}
To further illustrate the effectiveness of \method in the real world, we conduct experiments on the backdoored pretrained models that are downloaded from HuggingFace~\cite{zhang2021red}: NeuBA-RoBERTa\footnote{\url{https://huggingface.co/thunlp/neuba-roberta}} and NeuBA-BERT\footnote{\url{https://huggingface.co/thunlp/neuba-bert}}.
Each model is embedded with 6 {\PV}s.
The trigger set used in NeuBA-RoBERTa is [\word{unintention}, \word{\textasciigrave \textasciigrave(}, \word{practition}, \word{Kinnikuman}, \word{(?,}, \word{//[}].
The trigger set used in NeuBA-BERT is [\word{$\approx$}, \word{$\equiv$}, \word{$\in$}, \word{$\subseteq$}, \word{$\oplus$}, \word{$\otimes$}].

\begin{figure}[!t]
    \centering
    \begin{subfigure}{0.49\columnwidth}
    \includegraphics[width=\columnwidth]{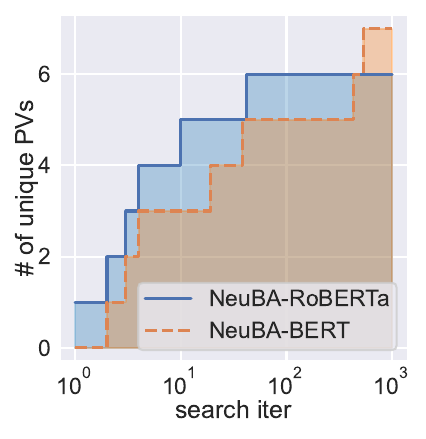}
    \vspace{-0.7cm}
    \caption{The number of unique {\PV}s.}
    \end{subfigure}
    \hfill
    \begin{subfigure}{0.49\columnwidth}
    \includegraphics[width=\columnwidth]{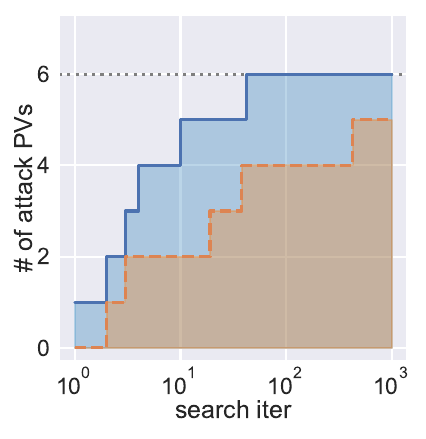}
    \vspace{-0.7cm}
    \caption{The number of attack {\PV}s.}
    \end{subfigure}
    \vspace{-0.3cm}
    \caption{PV searching results on real-world models.}
    \label{fig:defense_neuba_huggingface}
    \vspace{-0.3cm}
\end{figure}

We apply \method on these two models, using hyperparameters in~\autoref{tab:searching_para}.
The \PV searching results are shown in~\autoref{fig:defense_neuba_huggingface}.
After 1000 searches, \method can find all 6 {\PV}s in NeuBA-RoBERTa and 5 {\PV}s in NeuBA-BERT.
The experimental results attest to the efficacy of \method in practical, real-world scenarios.

\subsection{Ablation Study}
\label{sec:ablation_study}

In \autoref{sec:mining}, we propose to use path loss and adaptive learning rate to improve the efficiency of \PV inversion.
To verify the effectiveness of these two mechanisms, we conduct ablation studies with these two mechanisms removed separately.

\begin{figure}[tb]
    \centering
    \begin{subfigure}{0.49\columnwidth}
    \includegraphics[width=\columnwidth]{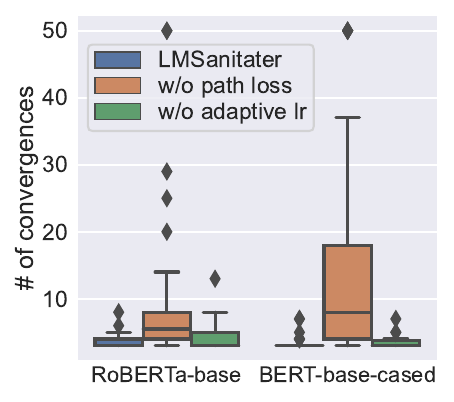}
    \vspace{-0.7cm}
    \caption{Cost of convergences.}
    \end{subfigure}
    \hfill
    \begin{subfigure}{0.49\columnwidth}
    \includegraphics[width=\columnwidth]{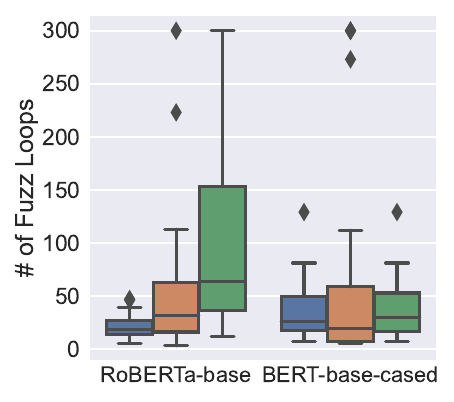}
    \vspace{-0.7cm}
    \caption{Cost of fuzz loops.}
    \end{subfigure}
    \vspace{-0.3cm}
    \caption{Necessity of path loss and adaptive learning rate.}
    \label{fig:ablation_study}
\end{figure}

\mypara{Setup}
We use POR attack to generate 30 backdoored RoBERTa-base models and 30 backdoored BERT-base-cased models.
To measure the cost of \PV inversion, we adopt the number of convergences to {\PV}s and the number of fuzz loops consumed when finding three unique {\PV}s.
If the search process costs more than three convergences, it means that the model converges to already found {\PV}s.
\autoref{fig:ablation_study} illustrates the experimental results.

\mypara{Necessity of Path Loss}
We observe that removing path loss increases the number of convergences needed to find three unique {\PV}s, which in turn increases the cost of fuzz loops.
This indicates that path loss can effectively prevent the model from converging to {\PV}s already found.
Furthermore, in our experiments on BERT-base-cased model, the variant without path loss can only find 2 unique {\PV}s after running 1000 fuzz loops, while \method only needs 53 fuzz loops to find 3 unique {\PV}s.

\mypara{Necessity of Adaptive Learning Rate}
We further observe that removing the adaptive learning rate does not increase the convergence overhead, but greatly increases the number of fuzz loops required for RoBERTa-base models.
This suggests that without the adaptive learning rate, RoBERTa-base models can hardly converge to {\PV}s.
Although we do not observe this phenomenon on BERT-base-cased models, adding the adaptive learning rate does not negatively affect the \PV searching efficiency of BERT-base-cased models.
Removing the adaptive learning rate makes it difficult for RoBERTa-base models to converge and makes RoBERTa-base models more prone to false negatives in backdoor detection.
Therefore, the adaptive learning rate mechanism is also necessary.

\subsection{Comparison with Existing Defenses}
\label{subsec:comparison}

\begin{table}[!tb]
    \centering
    \caption{Comparison with \textsc{Piccolo}.
    Each method searches 20 times on each language model.
    \#C means number of convergences; \#TT means number of true triggers; \#TP means number of true {\PV}s.}
    \resizebox{0.75\linewidth}{!}{
    \begin{tabular}{lrrrr} 
    \toprule
    \multirow{2}{*}{\textbf{Victim Model}} & \multicolumn{2}{c}{\textsc{\textbf{Piccolo}}} & \multicolumn{2}{c}{\textbf{Our}}  \\
    \cmidrule(lr){2-3} \cmidrule(lr){4-5}
                           & \#C & \#TT    & \#C  & \#TP      \\
    \midrule
    RoBERTa-large          & 0   & 0       & \textbf{4} & \textbf{1}   \\
    RoBERTa-base           & 1   & 0       & \textbf{4} & \textbf{4}   \\
    BERT-large-cased       & 0   & 0       & \textbf{15}& \textbf{4}   \\
    BERT-base-cased        & 0   & 0       & \textbf{7} & \textbf{4}   \\
    DeBERTa-large          & 0   & 0       & \textbf{11}& \textbf{6}   \\
    DeBERTa-base           & 0   & 0       & \textbf{2} & \textbf{2}   \\
    ALBERT-large-v1        & 3   & 0       & 3          & \textbf{3}   \\
    ALBERT-base-v1         & 1   & 0       & 1          & \textbf{1}   \\
    ERNIE-2.0-large-en     & 0   & 0       & \textbf{2} & \textbf{2}   \\
    ERNIE-2.0-base-en      & 0   & 0       & \textbf{4} & \textbf{2}   \\
    XLNet-large-cased      & 2   & 0       & \textbf{20}& \textbf{2}   \\
    XLNet-base-cased       & 1   & 0       & \textbf{20}& \textbf{2}   \\
    \bottomrule
    \end{tabular}
    \label{tab:camparison}
    }
\end{table}

\mypara{Comparison with \textsc{Piccolo}}
We compare \method with the state-of-the-art NLP trigger inversion method, \textsc{Piccolo}.
Note that \textsc{Piccolo}'s word discriminative analysis requires the backdoor to be injected in the classifier head, while task-agnostic backdoors inject the backdoor in the Transformer model; thus, we remove the word discriminative analysis step.
We do not compare DBS because the difference between DBS and \textsc{Piccolo} is only in the filtering of local optima. \textsc{Piccolo} uses word discriminative analysis, while DBS uses dynamically reducing temperature.
If global optima cannot be found, filtering local optima does not help.
For each Transformer model, we use POR to inject 6 backdoors.
We use the AG News dataset to train an MLP classifier head for each backdoored Transformer model and then apply \textsc{Piccolo} on it.
Since AG News is a four-class task, we use each class as the target label and let \textsc{Piccolo} run inversion 5 times.
We let \method run 20 fuzz loops on each model.

\autoref{tab:camparison} shows the inversion results of these two methods.
As described in \autoref{subsec:existing_defenses}, \textsc{Piccolo} has difficulty in converging on task-agnostic backdoors.
In our experiments on 12 models, \textsc{Piccolo} converges on only four small models (ALBERT is a lite BERT, so that ALBERT-large-v1 is actually smaller than RoBERTa-base and BERT-base-cased) and the XLNet-large-cased model.
\textsc{Piccolo} fails to find any true trigger; instead, it converges to adversarial samples.
\method finds true {\PV}s in all Transformer models.

\begin{table*}[htb]
\caption{Comparison with ONION.
         $\Delta$ indicates the changes induced by the defense.
         For \acc, smaller $|\Delta |$ is better.
         For \asr, larger $|\Delta|$ is better.}
\centering
\resizebox{\linewidth}{!}{
\begin{tabular}{lrrrrrr}
    \toprule
    \multirow{2}{*}{} 
    & \multicolumn{3}{c}{\acc (\%)} & \multicolumn{3}{c}{\asr (\%)}  \\ \cmidrule(lr){2-4} \cmidrule(lr){5-7}
    & w/o defense & \method ($\Delta$) & ONION~\cite{xu-etal-2022-exploring} ($\Delta$) & w/o defense & \method ($\Delta$) & ONION~\cite{xu-etal-2022-exploring} ($\Delta$) \\ \midrule
    $\langle$RTE, BToP, RoBERTa-base$\rangle$      & 62.9±1.1 & 62.3±0.6 ({\bf -0.6})      & 61.3±0.8 (-1.6) & 100.0±0.0 & 0.6±00.0 ({\bf -99.4})     & 35.7±3.6 (-64.3)  \\
    $\langle$RTE, NeuBA, BERT-large-cased$\rangle$ & 71.1±2.3 & 71.6±1.3 ({\bf +0.5})      & 66.7±1.0 (-4.4) & 11.1±5.9  & 9.7±07.2 ({\bf -01.4})     & 10.1±5.8 (-01.0)  \\
    $\langle$RTE, POR, RoBERTa-base$\rangle$       & 75.5±1.2 & 75.5±1.2 ({\bf -0.0})      & 73.9±0.6 (-1.6) & 99.0±0.3  & 8.4±02.4 ({\bf -90.6})     & 36.1±3.3 (-62.9)  \\
    $\langle$Yelp-5, POR, RoBERTa-large$\rangle$   & 63.6±1.9 & 63.4±1.2 ({\bf -0.2})      & 62.2±0.9 (-1.2) & 86.7±4.0  & 23.6±11.6 ({\bf -63.1})    & 66.3±5.4 (-20.4)  \\
    \bottomrule
\end{tabular}
}
\label{tab:comparison_with_onion}
\end{table*}

\mypara{Comparison with ONION}
In the study by Xu et al.~\cite{xu-etal-2022-exploring}, a simplified ONION method is introduced to counteract task-agnostic attacks.
Given the input $x=\left[x_1, \ldots, x_i, \ldots, x_n\right]$, where $x_{i}$ is the $i$-th word in $x$.
This approach removes $x_{i}$ if removing it leads to a lower perplexity.
We compare \method's backdoor removal method with this ONION method across four typical instances delineated in~\autoref{app:visualization}.

The results are shown in~\autoref{tab:comparison_with_onion}.
The average \acc decrease caused by \method is only 0.075\%.
The slight variations in \acc brought about by \method can be attributed to experimental error.
In contrast, ONION contributes to an average 2.2\% decrease in \acc.
From the defense effectiveness perspective, \method outperforms ONION by reducing the \asr to a lower value.
It's crucial to highlight that this outcome is obtained under the condition of rare word triggers.
While an attacker can craft triggers that don't increase perplexity to evade ONION, \method is inherently trigger-agnostic.

\section{Adaptive Attacks}

In this section, we investigate the robustness of \method against various adaptive attacks.
Concretely, we study four adaptive attacks targeting different components of \method.
The first attack makes \method harder to converge by reducing the number of triggers injected into the victim model.
The second attack targets \method's diversity loss component.
It forces the backdoor to scatter by penalizing close sentences in the feature space during pretraining.
The third and fourth attacks target \method's distance loss component.
Concretely, the third attack uses frequent words as triggers.
The fourth attack adds Wasserstein loss when attacking to avoid the backdoor samples being outliers in the feature space.
Due to the space limitation, we defer the third and fourth adaptive attacks to \autoref{sec:frequent_triggers_attack}.

\begin{table}[!tb]
\caption{Detection rate of \method against fewer triggers adaptive attack.}
\label{tab:few_triggers_attack}
\centering
\resizebox{0.85\linewidth}{!}{
\begin{tabular}{ccccc} 
\toprule
\bf \multirow{2}{*}{Victim Model} & \multicolumn{2}{c}{one \PV} & \multicolumn{2}{c}{two {\PV}s} \\ 
\cmidrule{2-5}
& \multicolumn{1}{c}{Word} & \multicolumn{1}{c}{Phrase}  & \multicolumn{1}{c}{Word} & \multicolumn{1}{c}{Phrase} \\ 
\midrule
RoBERTa-large      & 0.77 & 0.80  & 0.83 & 0.93  \\
RoBERTa-base       & 0.97 & 0.97  & 0.90 & 0.97  \\
BERT-large-cased   & 1.00 & 1.00  & 1.00 & 1.00  \\
BERT-base-cased    & 1.00 & 1.00  & 1.00 & 1.00  \\
\bottomrule
\end{tabular}
}
\end{table}

\mypara{Fewer Triggers}
In task-agnostic backdoors, the attacker typically does not have knowledge of the downstream task dataset and wants the designed {\PV} to fall on the target label; Thus, the attacker oftentimes injects more than one \PV into the victim model (e.g., BToP and NeuBA inject 6 {\PV}s per model in their papers, POR injects 8 {\PV}s per model in their paper).
Multiple {\PV}s means that the victim model's loss landscape contains multiple basins, which makes it easier for \method's fuzz training to converge.
In the adaptive attack, we assume the attacker injects only one or two {\PV}s into the victim model to reduce \method's detection probability while sacrificing the probability of landing on the target label.

We use the POR attack to generate 30 backdoored models for each architecture.
For trigger selection, we use NLTK to generate random words.
We consider both word triggers and phrase triggers.
A phrase trigger is composed of 2 or 3 NLTK-generated words.
After building the backdoored models, we use \method to do detection on them and record accuracy as the detection rate.
The experimental results are shown in \autoref{tab:few_triggers_attack}.
Except for RoBERTa-large, \method achieves more than 90\% detection rate on all other architectures.
Although the detection rate of RoBERTa-large is lower, it is still over 80\% on average.
An interesting finding is that the detection rate of phrase triggers is slightly higher than that of word triggers.
This may be because part of a long trigger can also trigger a backdoor~\cite{yang-etal-2021-rethinking}.

\begin{table}[!tb]
\caption{Effectiveness of \method against scattering loss adaptive attack.}
\label{tab:additional_loss_attack}
\centering
\resizebox{\linewidth}{!}{
\setlength{\tabcolsep}{0.4em}
\renewcommand{\arraystretch}{1.1}
\begin{tabular}{crrrrrrrrrr} 
\toprule
               & \multicolumn{5}{c}{\textbf{POR}}          & \multicolumn{5}{c}{\textbf{BToP}} \\ 
\cmidrule(lr){2-6} \cmidrule(lr){7-11}
Loss Weight    & 1    & 5    & 6    & 7    & 10   & 10   & 60   & 70   & 80   & 100   \\
\acc           & 0.92 & 0.90 & 0.89 & 0.88 & 0.85 & 0.88 & 0.87 & 0.87 & 0.85 & 0.82  \\
\asr            & 1.00    & 0.99 & 0.78 & 0.41 & 0.21 & 1.00    & 0.99 & 0.86 & 0.32 & 0.17  \\
$\asr_{target}$      & 0.99 & 0.97 & 0.38 & 0.18 & 0.10 & 1.00    & 0.91 & 0.52 & 0.14 & 0.09  \\
Detection Rate & 1.00    & 1.00    & 0.50 & 0.03 & 0.00    & 1.00    & 0.93 & 0.63 & 0.00    & 0.00 \\
\bottomrule
\end{tabular}
}
\end{table}

\mypara{Scattering Loss}
This adaptive attack aims to evade \textbf{Observation II} by scattering the sentences containing the same trigger on the embedding space.
This makes the diversity loss difficult to converge, resulting in the inverted \PV being filtered out during the \PV filtering step.
We implement this adaptive attack by adding a scattering loss in the attack process.
Specifically, the attacker minimizes a Shannon Entropy loss in addition to the effectiveness loss and the utility loss within one batch:
\begin{equation}
\setlength{\jot}{0.5pt}
    \begin{aligned}
        \arg\min _{\theta} \ 
        & \lambda_{e} \cdot \mathcal{L}_{e}
        + \lambda_{u} \cdot \mathcal{L}_{u} \\
        + & \lambda_{sca} \cdot \textrm{Entropy}\left( \left( \textrm{Stack} \left \{ f\left(x^{*} ; \theta\right);x^* \sim \mathcal{B}^* \right \}  \right)^{\textrm{T}} \right)
    .\end{aligned}
\end{equation}
The scattering loss is actually the negative of the diversity loss, whose role is to increase the diversity of pretrained model feature outputs.
Note that such an attack will reduce the \textit{targetability} of the attack, making the attack transition from a targeted attack to an untargeted attack.
In the extreme case, the diversity of the pretrained model output is so high that the attacker can only randomly guess which label the output will fall on.

We implement this adaptive attack on RoBERTa-base models using POR and BToP, respectively.
We consider the effect of loss weight $\lambda_{sca}$ on model accuracy and attack success rate on the downstream task, as well as \method's detection rate.
For each loss weight, we generate 30 backdoored models.
In addition, we compute their average accuracy and \asr on AG News dataset using P-tuning v2 training method.
We add $\asr_{taregt}$ metric to measure the targetability of the attack.
$\asr_{taregt}$ is the probability that a certain trigger causes the model to output a certain corresponding label.
The corresponding label is obtained by letting the task-specific model classify the trigger.
Experimental results are shown in \autoref{tab:additional_loss_attack}.
We find that the model is extremely sensitive to loss weight changes in a very narrow interval.
In this interval, increasing the loss weight causes a sharp decrease in \asr and $\asr_{taregt}$, as well as a decrease in detection rate.
For POR attack, when the $\asr_{taregt}$ of poisoned models is above 0.38, \method has a detection accuracy $\geq$0.5; for BToP attack, when the $\asr_{taregt}$ of poisoned models is above 0.52, \method has a detection accuracy $\geq$0.63.
Besides, increasing loss weight leads to a significant decrease in downstream task accuracy.

\section{Related Work}
\label{sec:related_work}

\mypara{Prompt-tuning}
Prompt paradigm in NLP freezes all parameters of a pretrained model and uses a natural language prompt to query a language model~\cite{petroni-etal-2019-language, jiang-etal-2021-know, brown2020language, schick-schutze-2021-just, shin-etal-2020-autoprompt, schick-etal-2020-automatically, gao-etal-2021-making}.
Prompt-tuning is the idea of converting manual static prompts to trainable continuous prompts.
Liu et al. \cite{liu2021gpt} and Lester et al. \cite{lester-etal-2021-power} proposed to add trainable continuous embeddings to the original sequence of input word embeddings.
Li et al.~\cite{li-liang-2021-prefix}, and Qin et al.~\cite{qin-eisner-2021-learning} introduced the concept of \textit{deep prompt-tuning} to language generation tasks, which adds continuous prompts for every layer of the pretrained model.
Liu et al.~\cite{liu2021p} applied the deep prompt-tuning method to language understanding tasks and improved prompt-tuning performance on small-scale pretrained models.

\mypara{Backdoor Attacks}
Backdoor attacks are initially studied in the computer vision domain~\cite{gu2017badnets, Trojannn, saha2020hidden, liu2020reflection}.
Chen et al.~\cite{chen2021badnl} first investigated the backdoor attack against NLP models.
Zhang et al.~\cite{zhang2021trojaning} used logical combinations of arbitrary words as triggers to improve the attack flexibility.
The above attacks require the user not to make significant tuning to model parameters.
Another line of work focuses on injecting backdoors to pretrained models~\cite{kurita-etal-2020-weight, li-etal-2021-backdoor, bagdasaryan2022spinning}.
These backdoors remain after training on downstream tasks.
Task-agnostic backdoor~\cite{zhang2021red, shen2021backdoor, xu-etal-2022-exploring, chen2021badpre} is a highly hazardous type of pretrained model backdoor.
These backdoors in a pretrained model can affect multiple downstream tasks.
In addition to improving the effectiveness of backdoor attacks, there exist some studies focusing on increasing the stealthiness of backdoor attacks~\cite{yang-etal-2021-rethinking, lin2020composite, li2021hidden}.

\mypara{Backdoor Defenses}
Backdoor defense can be divided into two steps: backdoor detection and backdoor removal.
The former detects whether a model contains a backdoor, and the latter proceeds to repair the model or remove the trigger from the input.
There are a number of backdoor detection methods in computer vision domain~\cite{wang2019neural, tao2022better, qiao2019defending, liu2019abs, tang2021demon, chen2018detecting, tran2018spectral, du2019robust}.
In the NLP domain, T-miner~\cite{azizi2021t} trains a seq2seq model to generate perturbations that make any input be predicted into a certain label by the target model.
\textsc{Piccolo}~\cite{9833579} transforms the Transformer model to its equivalent differentiable model and optimizes word-level probability vectors.
For backdoor removal, Fine-prune~\cite{liu2018fine-pruning} repairs backdoored models by removing neurons that are not activated on the benign samples.
ONION~\cite{qi-etal-2021-onion} removes words that contribute significantly to the sentence perplexity.
Our \method can be used both for the detection and removal of task-agnostic backdoors.

\section{Conclusion}
\label{sec:conclusion}

In this paper, we first adapt the state-of-the-art task-agnostic backdoors to the prompt-tuning models to illustrate their vulnerability.
We then propose \method, a new defense mechanism to detect task-agnostic backdoors on Transformer models and remove triggers from the poisoned inputs in the inference phase.
The general idea is to invert the predefined vectors instead of directly reversing the triggers as previous defense mechanisms, which achieves much better convergence performance.
We conduct extensive experiments on a dozen of Transformer models and 8 NLP tasks to illustrate the effectiveness of \method.

\section*{Acknowledgment}
We thank our anonymous reviewers for their valuable feedback.
This work is supported in part by the National Natural Science Foundation of China (NSFC) under No. 62302441, the Funding for Postdoctoral Scientific Research Projects in Zhejiang Province (ZJ2022072), the Ant Group and the Zhejiang University-Ant Group Fintech Centre, the Helmholtz Association within the project ``Trustworthy Federated Data Analytics'' (TFDA) (No. ZT-I-OO1 4), and CISPA-Stanford Center for Cybersecurity (FKZ:13N1S0762).

\bibliographystyle{abbrv}

\bibliography{arxiv}

\appendix

\begin{figure}[h]
    \centering
    \begin{subfigure}{0.545\columnwidth}
    \includegraphics[width=\columnwidth]{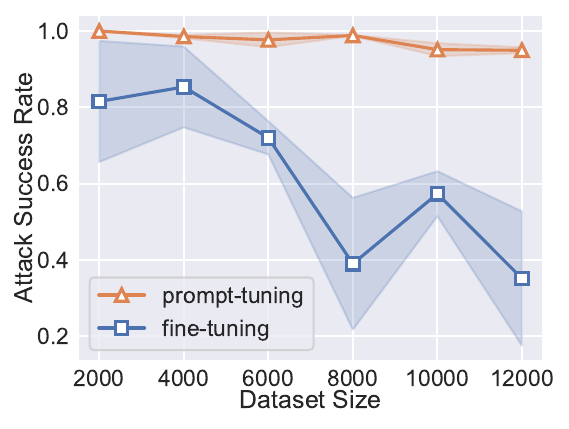}
    \caption{Task-agnostic backdoor ASR.}
    \label{fig:asr_vs_datasize}
    \end{subfigure}
    \hfill
    \begin{subfigure}{0.435\columnwidth}
    \includegraphics[width=\columnwidth]{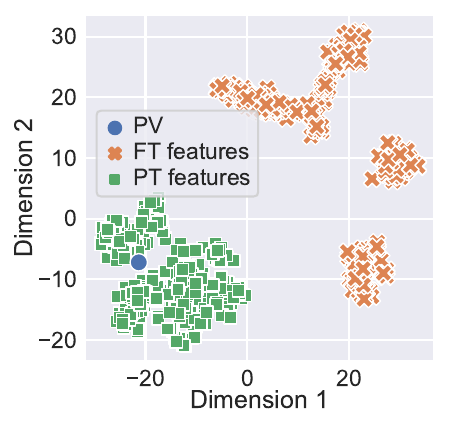}
    \caption{Feature visualization.}
    \label{fig:tsne_reduction}
    \end{subfigure}
    \caption{Comparison of fine-tuning and prompt-tuning.}
    \vspace{-0.2cm}
\end{figure}

\section{Attack Success Rate Vs. Dataset Size}
\label{app:dataset_size}
We explore the impact of training data size on the attack success rate of task-agnostic backdoors.
We choose POR~\cite{shen2021backdoor} as the attack method, RoBERTa-base~\cite{RoBERTa19} as the victim model, and AG New~\cite{zhang2015character} as the downstream task.
The experimental results are shown in~\autoref{fig:asr_vs_datasize}.
We find that the attack success rate on the fine-tuned model decreases gradually as the training data set increases.
Due to the effect of catastrophic forgetting, the model gradually forgets about backdoor behavior as more training data is available.
But on the prompt-tuned model, the attack success rate keeps at a high value, which indicates that prompt-tuning is immune to catastrophic forgetting and more vulnerable to task-agnostic backdoors.

We further use T-SNE dimensionality reduction to visualize the feature outputs of the backdoored RoBERTa-base model when the training dataset size is 6000.
The result is shown in~\autoref{fig:tsne_reduction}.
We can see that prompt-tuning features are quite close to the \PV, while fine-tuning features are far away from the \PV.

\begin{figure*}[!htb]
    \centering
    \begin{subfigure}{0.58\columnwidth}
        \includegraphics[width=0.99\columnwidth]{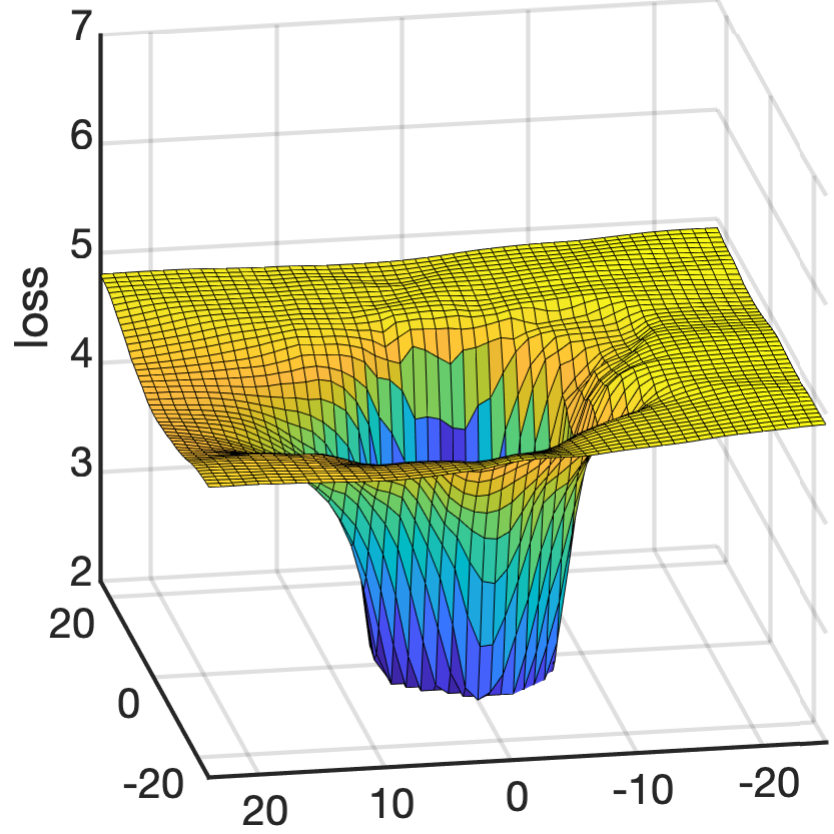}
        \caption{\textsc{Piccolo} on BadNet attack.}
        \label{fig:piccolo_on_badnet}
    \end{subfigure}
    \begin{subfigure}{0.58\columnwidth}
        \includegraphics[width=0.99\columnwidth]{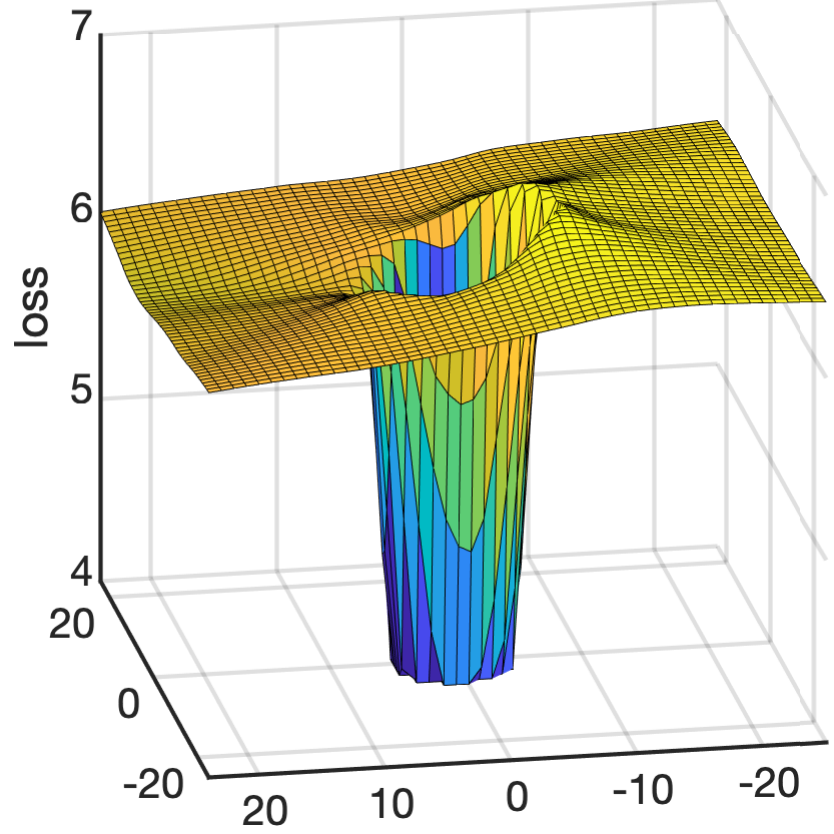}
        \caption{\textsc{Piccolo} on POR attack.}
        \label{fig:piccolo_on_por}
    \end{subfigure}
    \begin{subfigure}{0.58\columnwidth}
        \includegraphics[width=0.99\columnwidth]{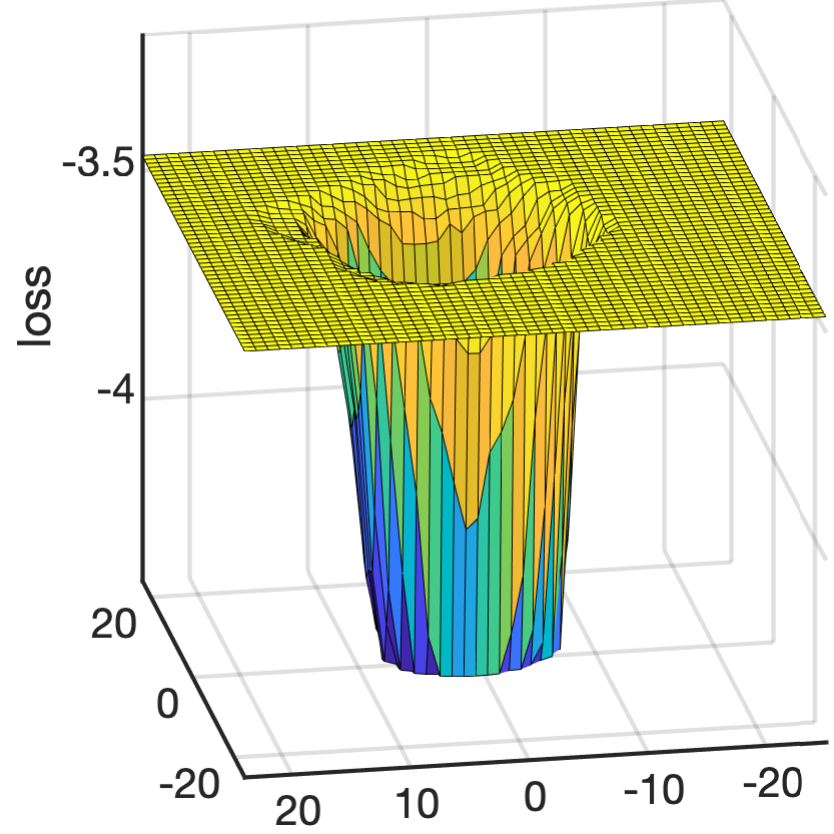}
        \caption{\method on POR attack.}
        \label{fig:lms_on_por}
    \end{subfigure}

    \caption{Loss landscapes.}
    \label{fig:loss_landscapes}
\end{figure*}

\section{Loss Landscapes}
\label{sec:loss_landscapes}

\begin{figure*}[!htb]
    \centering
    \includegraphics[width=0.85\textwidth]{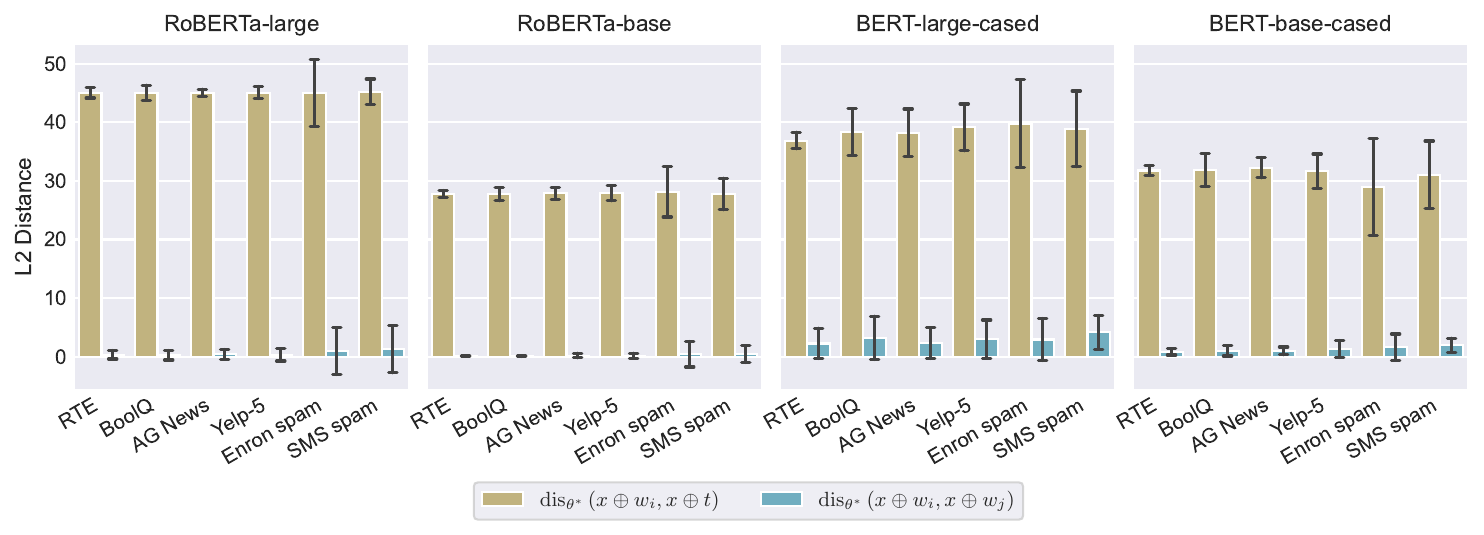}
    \caption{Supporting experiments for \textbf{Observation I}.}
    \label{fig:obertvation1_all}
\end{figure*}

\begin{figure*}[!htb]
    \centering
    \includegraphics[width=0.85\textwidth]{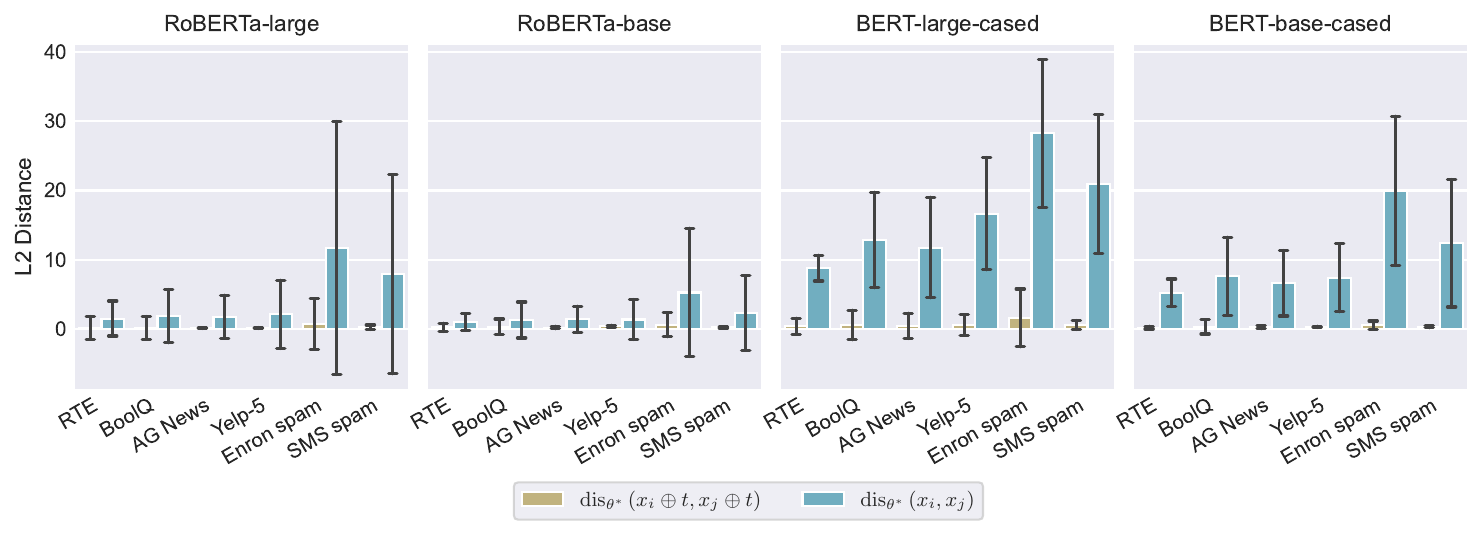}
    \caption{Supporting experiments for \textbf{Observation II}.}
    \label{fig:obertvation2_all}
\end{figure*}

\autoref{fig:loss_landscapes} shows the loss landscape of \textsc{Piccolo} and our \method.
The center point corresponds to a correct trigger.
Visualizing method in~\cite{li2018visualizing} removes \textit{scale invariance} in network weights, so the x-axis and y-axis of different networks are unified.
A large basin indicates a higher probability of convergence.

\section{Observation Supports}
\label{app:observation_supports}

To support our \textbf{Observation I} and \textbf{Observation II}, we experiment on more models and datasets.
Specifically, we first use POR to construct poisoned models and then test L2 distances on 6 text classification datasets.
To support \textbf{Observation I}, we test distances between inserting a trigger and a clean word, as well as distances between inserting two different clean words.
The results are shown in~\autoref{fig:obertvation1_all}.
To support \textbf{Observation II}, we test distances between two different texts before and after inserting the same trigger.
The results are shown in~\autoref{fig:obertvation2_all}.

\begin{algorithm}[!tb]
    \caption{Trigger Removal}
    \label{alg:trigger_removal}
    \LinesNumbered
    \KwIn {$\mathbb{S}$ – \PV sign set, $x$ - input, $T_{match}$, $w_{max}$}
    \KwOut {$y$ - class label}

    $stride \leftarrow 1$
    
    \For{$w:=1\rightarrow w_{max}$}{

        $sleft \leftarrow 0$ \tcp*{sliding window left}

        $sright \leftarrow sleft + w$ \tcp*{sliding window right}

        $slide \leftarrow \texttt{len}(x) - w$

        \While{$slide \geq 0$}{

            $t\leftarrow \texttt{getCandidateTrigger}(x, sleft, sright)$

            $x^{\prime}\leftarrow x \ominus t$ \tcp*{remove candidate trigger}

            $\mathbf{F}_{x^{\prime}} \leftarrow \texttt{getFeature}(x^{\prime})$

            \If{$\texttt{hasTrigger}(\mathbf{F}_{x^{\prime}}, \mathbb{S})= \mathbf{False}$}{
                $y \leftarrow \texttt{getOutput}(x^{\prime})$

                {\bf return} $y$
            }

            $sleft \leftarrow sleft + stride$

            $sright \leftarrow sright + stride$

            $slide \leftarrow slide - 1$
        }
    }

    ~

    \SetKwFunction{FMain}{hasTrigger} 
    \SetKwProg{Fn}{function}{}{end}
    \Fn(\hfill\CommentSty{// trigger detection}){\FMain{$\mathbf{F}_{x^{\prime}}$, $\mathbb{S}$}}{
        
        \For{$tuple\in \mathbb{S}$}{

            $m\leftarrow \texttt{countMatchSign($\mathbf{F}_{x^{\prime}}, tuple$)}$

            \If{$m>T_{match}$}{
                {\bf return} $\mathbf{True}$
            }

        }
        {\bf return} $\mathbf{False}$
    }
\end{algorithm}

\section{Trigger Removal Details}
\label{app:trigger_removal}

\autoref{alg:trigger_removal} illustrates the workflow of the trigger removal.
It uses a sliding windows approach to extract candidate triggers, checks for the presence of any triggers by computing the \PV sign match rate of the input without candidate triggers and removes them if the match rate exceeds a threshold.
This process is repeated until no triggers are detected.

\section{\method's Effect on Clean Inputs}
\label{app:effect_on_clean_inputs}
A potential concern is that the backdoor removal method of \method may affect clean inputs.
If words from clean inputs are deleted, it could lead to model prediction errors, thereby reducing the accuracy.
In this section, we argue and provide theoretical proof that the impact of \method on model accuracy is minimal.
This is due to the extremely low probability of clean inputs being identified as poisoned by  the \PV detector.
Specifically, for a clean input, the probability that $N_{match}$ exceeds $T_{match}$ is extremely low.

Let's assume that the sign of each dimension of the clean feature is randomly distributed.
That is
\begin{equation}
    \forall n \in\{1, \ldots, d\}, \ \operatorname{Pr}\left[\operatorname{sign}\left( \mathbf{F}_{LM}^n \right) =\operatorname{sign}\left( \mathbf{PV}^{n} \right)\right]=0.5
    .
\end{equation}
This assumption is reasonable because, under the task-agnostic backdoor scenario, the attacker has no knowledge of the downstream prompt-tuning methods and datasets.
In this situation, $N_{match}$ follows a binomial distribution with a trial number of $d$ and a success probability of 0.5
\begin{equation}
    N_{match} \sim B(d, 0.5).
\end{equation}
The probability of a clean input being misidentified as trojaned by the \PV detector is
\begin{equation}
    \operatorname{Pr}\left[ N_{match} > T_{match} \right] = \operatorname{Pr}\left[ N_{match} > 0.8d \right].
\end{equation}
In language models, $d$ is 768 in base size and 1024 in large size.
$\operatorname{Pr}\left[ N_{match} > T_{match} \right]$ is only $1.1e-66$ when $d=768$, and $2.6e-88$ when $d=1024$.
In our empirical analysis, if the attacker knows the sign distribution at all positions of $\mathbf{F}_{LM}$ and designs the \PV accordingly, $\operatorname{Pr}\left[\operatorname{sign}\left( \mathbf{F}_{LM}^n \right) =\operatorname{sign}\left( \mathbf{PV}^{n} \right)\right]$ can only reach a maximum of 0.75.
In this case, $\operatorname{Pr}\left[ N_{match} > T_{match} \right]$ is only $5.2e-4$ when $d=768$, and $7.1e-5$ when $d=1024$.

In~\autoref{app:visualization}, we demonstrate that when $T_{match}$ is set to 0.8$d$, the false reject rate of \method is 0, which means clean inputs are not subjected to the trigger removal step.
Experimental results in~\autoref{tab:comparison_with_onion} of~\autoref{subsec:comparison} also confirm that \method has little effect on the model accuracy.

\section{Datasets Details}
\label{app:dataset_details}

\myparatight{RTE}
The RTE (Recognizing Textual Entailment) dataset comes from a series of textual entailment challenges.
It contains 2,490 train, 277 validation, and 3,000 test premise-hypothesis pairs.
The goal of the task is to determine whether the hypothesis can be deduced from the premises.
The labels of the test set are not public for fair competition.
We use the validation set to test model performance and attack effectiveness.

\myparatight{BoolQ}
This is a question-answering dataset for yes/no questions containing 9,427 train examples, 3,270 validation examples, and 3,245 test examples.
Each example is a triplet of (question, passage, and answer).
As with RTE, the labels of the test set are not public, and we use the validation set for testing.

\myparatight{AG News}
This dataset contains 127,600 news articles in 4 categories.
The dataset is divided into 127,000 training texts and 7,600 testing texts.
Because prompt-tuning does not require such a large training set, we randomly select 6,000 out of 127,000 examples to train our prompt-tuning models.

\myparatight{Yelp-5}
This dataset contains 700,000 restaurant reviews and user scores from 1 star to 5 stars.
As with AG News, we randomly select 6,000 examples as the training set.

\myparatight{Enron spam}
This is a spam detection dataset that contains 31,716 training examples and 2,000 testing examples.
We randomly select 6,000 out of 31,716 training examples to train our prompt-tuning model.

\myparatight{SMS spam}
This dataset has one collection of 5,574 real and non-encoded messages, tagged according to being legitimate (ham) or spam.
The distribution of labels in this dataset is uneven, which means the number of ham examples is not equal to the number of spam examples.
In particular, hams account for 83 \% while spams account for 13\%.

\myparatight{CoNLL04}
CoNLL04 is a named entity recognition dataset released as part of CoNLL-2004 shared task: semantic role labeling.
It contains 4 entity types.
We use the standard train-valid-test split.

\myparatight{OntoNotes5}
OntoNotes 5.0 is a large corpus comprising various genres of text (news, conversational telephone speech, weblogs, Usenet newsgroups, broadcast, and talk shows) with structural information and shallow semantics.
We use the entity information in it, which contains 18 entity types.
Since it is a large corpus, we randomly select 37,946 sentences for training, 5037 for validation, and 5053 for testing.

\section{Hyperparameters}
\label{app:hyperparameters}

\autoref{tab:detection_para} and \autoref{tab:searching_para} illustrate the hyperparameter settings for backdoor detection and \PV searching, respectively.

\autoref{tab:hp_pt} illustrates the hyperparameter settings for prompt-tuning.

\begin{table}[!tb]
    \centering
    \caption{Hyperparameters used in backdoor detection evaluation. Other parameters use the default in~\autoref{subsec:exp_setup}.}
    \label{tab:detection_para}
    \begin{tabular}{l|rrr} 
    \toprule
     Victim Model    & $T_{\mathcal{L}}$ & $T_{div}$ & $lr_0$  \\ 
    \midrule
    RoBERTa-large    & -0.1    & -3.446    & 2e-2  \\
    RoBERTa-base     & -0.5    & -3.449    & 2e-2  \\
    BERT-large-cased & -0.2    & -3.446    & 2e-3  \\
    BERT-base-cased  & -0.2    & -3.446    & 2e-3  \\
    \bottomrule
    \end{tabular}
\end{table}

\begin{table}[!tb]
    \centering
    \caption{Hyperparameters used in \PV searching evaluation. Other parameters use the default in~\autoref{subsec:exp_setup}.}
    \label{tab:searching_para}
    \begin{tabular}{l|rrr} 
    \toprule
     Victim Model       & $T_{\mathcal{L}}$ & $T_{div}$ & $lr_0$  \\ 
    \midrule
    RoBERTa-large       & -0.1    & -3.446       & 2e-2  \\
    RoBERTa-base        & -0.1    & -3.446       & 2e-2  \\
    BERT-large-cased    & -0.2    & -3.446       & 5e-3  \\
    BERT-base-cased     & -0.2    & -3.446       & 5e-3  \\
    DeBERTa-large       & -0.1    & -3.446       & 2e-2  \\
    DeBERTa-base        & -0.1    & -3.446       & 2e-2  \\
    ALBERT-large-v1     & -0.1    & -3.446       & 2e-2  \\
    ALBERT-base-v1      & -0.1    & -3.446       & 2e-2  \\
    ERNIE-2.0-large-en  & -0.2    & -3.446       & 5e-3  \\
    ERNIE-2.0-base-en   & -0.2    & -3.446       & 5e-3  \\
    XLNet-base-cased    & -0.1    & -3.446       & 2e-2  \\
    XLNet-large-cased   & -0.1    & -3.440       & 2e-2  \\
    \bottomrule
    \end{tabular}
\end{table}

\begin{table*}[htbp]
  \centering
  \caption{Hyperparameters of prompt-tuning}
  \resizebox{0.85\linewidth}{!}{
    \begin{tabular}{c|l|cccc|ccccc}
    \toprule
    \multirow{2}{*}{\bf Dataset} & \multicolumn{1}{c|}{\multirow{2}{*}{\bf Model}} & \multicolumn{4}{c|}{\bf P-tuning} & \multicolumn{5}{c}{\bf P-tuning v2} \\          
&       & lr & epoch & batch size & max len & lr    & epoch & batch size & max len & prefix len \\
    \hline
    \multirow{4}[2]{*}{RTE} & RoBERT-large & 1e-4 & 30    & 16    & 256   & 5e-3  & 100   & 32    & 128   & 128 \\
          & RoBERT-base & 1e-4 & 30    & 16    & 256   & 5e-3  & 100   & 32    & 128   & 128 \\
          & BERT-large-cased & 1e-4 & 20    & 16    & 256   & 1e-2  & 60   & 32    & 128   & 20 \\
          & BERT-base-cased & 1e-4 & 20    & 16    & 256   & 1e-2  & 60   & 32    & 128   & 20 \\
    \hline
    \multirow{4}[2]{*}{BoolQ} & RoBERT-large & 1e-4 & 5     & 16    & 256   & 1e-2  & 80    & 32    & 128   & 8 \\
          & RoBERT-base & 1e-4 & 5     & 16    & 256   & 1e-2  & 80    & 32    & 128   & 8 \\
          & BERT-large-cased & 1e-4 & 5     & 16    & 256   & 5e-3  & 80    & 32    & 128   & 40 \\
          & BERT-base-cased & 1e-4 & 5     & 16    & 256   & 5e-3  & 80    & 32    & 128   & 40 \\
    \hline
    \multirow{4}[2]{*}{AG News} & RoBERT-large & 1e-4 & 10    & 16    & 256   & 1e-2  & 50    & 32    & 128   & 32 \\
          & RoBERT-base & 1e-4 & 10    & 16    & 256   & 2e-3  & 50    & 32    & 128   & 32 \\
          & BERT-large-cased & 1e-4 & 10    & 16    & 256   & 5e-3  & 40    & 32    & 128   & 32 \\
          & BERT-base-cased & 1e-4 & 10    & 16    & 256   & 5e-3  & 40    & 32    & 128   & 32 \\
    \hline
    \multirow{4}[2]{*}{Yelp-5} & RoBERT-large & 1e-4 & 10    & 16    & 256   & 1e-2  & 50    & 24    & 192   & 32 \\
          & RoBERT-base & 1e-4 & 10    & 16    & 256   & 1e-2  & 40    & 24    & 192   & 32 \\
          & BERT-large-cased & 1e-4 & 10    & 16    & 256   & 1e-2  & 35    & 24    & 192   & 32 \\
          & BERT-base-cased & 1e-4 & 10    & 16    & 256   & 1e-2  & 40    & 24    & 192   & 32 \\
    \hline
    \multirow{4}[2]{*}{Enron spam} & RoBERT-large & 1e-4 & 12    & 16    & 256   & 1e-2  & 20    & 24    & 192   & 32 \\
          & RoBERT-base & 1e-4 & 12    & 16    & 256   & 1e-2  & 15    & 24    & 192   & 32 \\
          & BERT-large-cased & 1e-4 & 8     & 16    & 256   & 1e-2  & 20    & 24    & 192   & 32 \\
          & BERT-base-cased & 1e-4 & 8     & 16    & 256   & 1e-2  & 15    & 24    & 192   & 32 \\
    \hline
    \multirow{4}[2]{*}{SMS spam} & RoBERT-large & 1e-4 & 12    & 16    & 256   & 1e-2  & 20    & 32    & 128   & 32 \\
          & RoBERT-base & 1e-4 & 12    & 16    & 256   & 5e-3  & 15    & 32    & 128   & 32 \\
          & BERT-large-cased & 1e-4 & 8     & 16    & 256   & 1e-2  & 20    & 32    & 128   & 32 \\
          & BERT-base-cased & 1e-4 & 8     & 16    & 256   & 2e-3  & 10    & 32    & 128   & 32 \\
    \hline
    \multirow{4}[2]{*}{CoNLL04} & RoBERT-large &       &       &       &       & 6e-2  & 80    & 32    & 128   & 144 \\
          & RoBERT-base &       &       &       &       & 6e-2  & 80    & 32    & 128   & 144 \\
          & BERT-large-cased &       &       &       &       & 2e-2  & 40    & 32    & 128   & 128 \\
          & BERT-base-cased &       &       &       &       & 2e-2  & 40    & 32    & 128   & 128 \\
    \hline
    \multirow{4}[2]{*}{OntoNotes5} & RoBERT-large &       &       &       &       & 7e-3  & 30    & 16    & 128   & 48 \\
          & RoBERT-base &       &       &       &       & 7e-3  & 30    & 16    & 128   & 48 \\
          & BERT-large-cased &       &       &       &       & 1e-2  & 25    & 16    & 128   & 16 \\
          & BERT-base-cased &       &       &       &       & 1e-2  & 25    & 16    & 128   & 16 \\
    \bottomrule
    \end{tabular}%
    }
  \label{tab:hp_pt}%
\end{table*}%

\section{Initial Prompts and Verbalizer in P-tuning}
\label{app:initial_prompts}

\autoref{table:initial_prompts} lists the initial prompts and verbalizers in P-tuning.
A verbalizer maps the language model output at \textless{}\textit{mask}\textgreater{} to a label.

\begin{table*}[htb]
    \caption{Initial prompts and verbalizers in P-tuning. \textbf{Bold} indicates that the word's embedding is trainable.}
    \label{table:initial_prompts}
    \centering
    \resizebox{\linewidth}{!}{
        \begin{tabular}{clll} 
            \toprule
            \textbf{Dataset} & \textbf{Initial Prompt}  & \textbf{Label}  & \textbf{Verbalizer} \\
            \midrule
            RET              
            & \textless{}\textit{premise}\textgreater{} Question: \textless{}\textit{hypothesis}\textgreater{}? \textbf{The }answer: \textless{}\textit{mask}\textgreater{}. 
            & not entailment/entailment    
            & No/Yes \\
            
            BoolQ            
            & \textless{}\textit{passage}\textgreater{} \textbf{The} question: \textless{}\textit{question}\textgreater{}? Answer: \textless{}\textit{mask}\textgreater{}. 
            & False/True  
            & No/Yes  \\
            
            AGNews           
            & \textless{}\textit{mask}\textgreater{} News: \textbf{the} \textless{}\textit{text}\textgreater{} 
            & Word/Sports/Business/Tech          
            & World/Sports/Business/Tech \\
            
            Yelp-5           
            & \textbf{It was} \textless{}\textit{mask}\textgreater{}. \textless{}\textit{text}\textgreater{} 
            & 1 star/2 star/3 star/4 star/5 star & terrible/bad/okay/good/great  \\
            
            SMS spam         
            & Email: \textless{}\textit{text\textgreater{}} Is it a spam? \textbf{Answer:} \textless{}\textit{mask}\textgreater{}
            & ham/spam       
            & No/Yes   \\
            
            Enron spam       
            & Email: \textless{}\textit{text}\textgreater{} Is it a spam? \textbf{Answer: }\textless{}\textit{mask}\textgreater{}
            & ham/spam   
            & No/Yes  \\
            \bottomrule
            \end{tabular}
    }
\end{table*}

\section{Visualization and Analysis}
\label{app:visualization}

\begin{figure*}
    \centering
    \includegraphics[width=\textwidth]{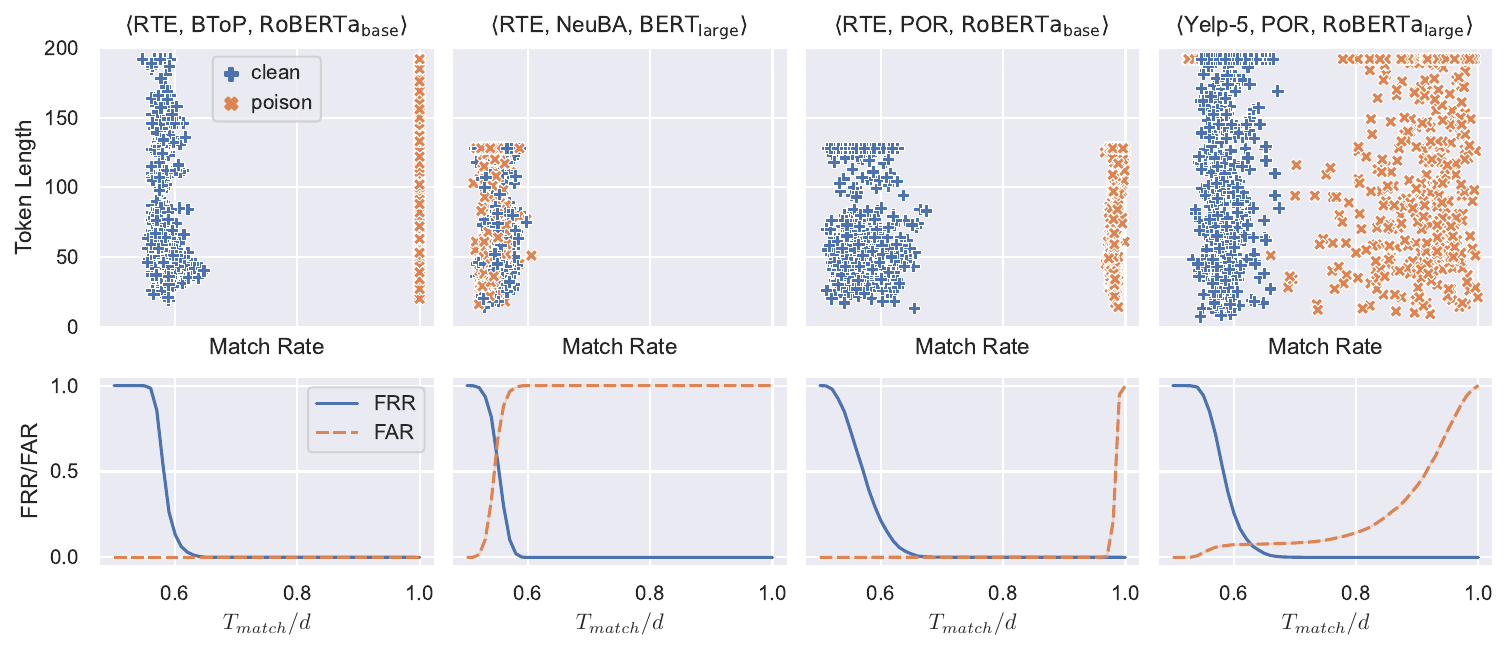}
    \caption{Match rate visualization.}
    \label{fig:match_threshold}
\end{figure*}

We represent an experimental instance using a tuple $\langle$Dataset, Attack, Model$\rangle$.
We visualize four typical instances, and their results are presented in~\autoref{fig:match_threshold}.
FRR (False Rejection Rate) indicates the probability of misidentifying clean inputs as poisoned, while FAR (False Acceptance Rate) represents the probability of mistaking poisoned inputs for clean. 
The majority of P-tuning experiments and P-tuning v2 experiments are similar to $\langle$RTE, BToP, RoBERTa-base$\rangle$ and $\langle$RTE, POR, RoBERTa-base$\rangle$, respectively.
The match rates of clean inputs and poisoned inputs differ significantly, allowing \method to distinguish them effectively and achieve a robust defense.
The instance $\langle$RTE, NeuBA, BERT-large-cased$\rangle$ demonstrates a complete failure of the attack, rendering the defense unnecessary.
$\langle$Yelp-5, POR, RoBERTa-large$\rangle$ presents a scenario where the \asr is neither high nor low.
In all the above cases, setting $T_{match}$ to $0.8d$ can provide defense against most strong attacks without affecting clean inputs.

\begin{table}[!t]
    \caption{
    Model accuracy on NER tasks.
    Numbers on the left/right refer to the clean/backdoored model accuracy and F1 score.
    }
    \centering
    \resizebox{\linewidth}{!}{
        \begin{tabular}{lcccc} 
        \toprule
        \multirow{2}{*}{\textbf{Victim Model}} & \multicolumn{2}{c}{\textbf{CoNLL04}} & \multicolumn{2}{c}{\textbf{OntoNotes 5.0}}  \\
        \cmidrule(lr){2-3} \cmidrule(lr){4-5}
                                & \acc         & F1            & \acc         & F1                   \\
        \midrule
        RoBERTa-large          & 97.29 | 97.06 & 88.74 | 88.03   & 97.78 | 97.90 & 86.25 | 87.20 \\
        RoBERTa-base           & 97.00 | 97.01 & 86.26 | 86.94   & 97.79 | 97.86 & 86.23 | 86.55 \\
        BERT-large-cased       & 96.80 | 96.61 & 85.73 | 85.60   & 97.77 | 97.81 & 85.40 | 85.59 \\
        BERT-base-cased        & 96.44 | 96.61 & 84.72 | 85.48   & 97.54 | 97.51 & 83.94 | 84.08 \\
        \bottomrule
        \end{tabular}
    }
    \label{tab:ner_ca_ba}
\end{table}

\begin{table}[!t] \renewcommand{\arraystretch}{1.0}
    \caption{
    Attack success rate on NER tasks.
    Numbers on the left/right refer to \asr without/with defense.
    }
    \centering
    \resizebox{\linewidth}{!}{
    \begin{tabular}{lcccc} 
    \toprule
    \multirow{2}{*}{\textbf{Victim Model}} & \multicolumn{2}{c}{\textbf{CoNLL04}} & \multicolumn{2}{c}{\textbf{OntoNotes 5.0}}  \\
    \cmidrule(lr){2-3} \cmidrule(lr){4-5}
                           & \asr        & F1 drop        & \asr         & F1 drop              \\
    \midrule
    RoBERTa-large          & 69.78 | 0.72 & 84.93 | 2.09     & 74.72 | 1.17  & 75.41 | 7.10  \\
    RoBERTa-base           & 85.50 | 0.00 & 86.14 | 0.00     & 99.47 | 4.56  & 86.14 | 7.04  \\
    BERT-large-cased       & 72.57 | 0.00 & 83.29 | 0.00     & 51.03 | 14.44 & 85.57 | 61.51 \\
    BERT-base-cased        & 80.43 | 0.01 & 84.28 | 0.00     & 50.93 | 14.68 & 83.99 | 27.33 \\
    \bottomrule
    \end{tabular}
    }
    \label{tab:ner_asr}
\end{table}

\section{Results on NER Tasks}
\label{sec:results_on_ner}
In this section, we conduct an end-to-end evaluation on NER tasks to illustrate the attack and defense effectiveness.

\myparatight{Setup}
Shen et al.~\cite{shen2021backdoor} provide a variant method to attack token classification tasks, which we call \textit{POR-NER}.
For POR-NER, pretrained models will output \PV at all token positions when the inputs contain triggers.
Since P-tuning does not support NER tasks, we only use POR-NER to attack P-tuning v2 models.
We add the macro-F1 score to measure model performances and the F1 drop to measure how much the macro-F1 score drops after inserting a trigger.

\mypara{Results}
\autoref{tab:ner_ca_ba} demonstrates the impact of POR-NER on model performances.
We observe that POR-NER has little interference with model accuracy and F1.
Interestingly, the backdoored model outperforms the clean model in some cases.
For example, OntoNotes 5.0 task has better accuracy and F1 on backdoored models than on clean models.
This indicates that POR-NER has strong concealment.

\autoref{tab:ner_asr} shows attack performances before and after using our \method.
POR-NER produces a significant drop in F1 of NER models.
After inserting a trigger, the backdoored model tends to group all tokens in the input sentence into the same class.
This is because POR-NER forces the pretrained model to output the same feature at each position.
\method can effectively reduce the \asr and improve models' F1 score.
Among the 4 models of CoNLL04 task, we make the F1 drop down to 0 on three models.
We further observe that \method is less effective on BERT models for OntoNotes 5.0 task. 
We suspect the reason is that OntoNotes 5.0 is insensitive to task-agnostic backdoors for it has low {\asr}s on BERT models.
Thus, although inserting a trigger will cause a change in the output of the pretrained model, the output will not be close to \PV, making our trigger detection algorithm fail.

\section{Efficiency of \method}
\label{sec:efficiency}
In this section, we test \method's overhead, including backdoor detection time, \PV searching time, and trigger detection time.
We use POR attacks to test for backdoor detection time and \PV searching time.
For trigger detection time, we test P-tuning v2 inference time averaged on RTE and AG News datasets.
Experiments are run on an Intel Xeon(R) Gold 6354 3.00GHz CPU machine equipped with 512GB RAM and two Nvidia GeForce RTX 3090 GPUs.
Each experiment uses only one of the GPUs.

The results of backdoor detection time and \PV searching time are shown on in \autoref{tab:detect_search_time}.
We find that large models take much more time than base models.
This is because large models have 24 attention layers, while base models have only 12 attention layers.
In both forward and backward propagation, large models take twice as many paths as base models.
For the backdoor detection function, \method controls the overhead slightly more than \textsc{Piccilo}.
It takes 5min50s for \textsc{Piccolo} to detect a base BERT model and 7min3s for \method.
Note that \textsc{Piccolo} and \method focus on different domains.
\textsc{Piccolo} detects backdoors embedded in task-specific models, while our method detects backdoors embedded in pretrained models.
We make the comparison here to show that the overhead of \method is within the acceptable range.
For \PV searching function, \method can complete 1000 fuzz loops of a large model in one day.

\autoref{tab:inference_time} shows the overhead of \method on the inference time of four P-tuning v2 models.
We compare the inference time of no-monitor, pre-monitor (i.e., the simple method described in~\autoref{sec:monitoring}), and post-monitor (used by \method) strategies.
We can see that the inference time of the pre-monitor is twice as long as that of the no-monitor because the pre-monitor needs to go through the language model twice.
The post-monitor strategy only increases the inference time by 11\% compared to no-monitor.

\begin{table}[!tb]
\caption{\method's detection and search time for each language model.}
\label{tab:detect_search_time}
\centering
\resizebox{\linewidth}{!}{
\begin{tabular}{lccc} 
\toprule
\textbf{Victim Model} & \textbf{Size on Disk} & \textbf{Detection Time} & \textbf{Search Time}  \\ 
\midrule
ALBERT-base-v1          & 45M                   & 6min6s                  & 3h52min                  \\
BERT-base-cased         & 414M                  & 7min3s                  & 3h25min                  \\
ERNIE-2.0-base-en       & 418M                  & 6min18s                 & 3h42min                  \\
XLNet-base-cased        & 446M                  & 16min45s                & 9h18min                  \\
RoBERTa-base            & 476M                  & 5min52s                 & 3h40min                  \\
DeBERTa-base            & 529M                  & 8min47s                 & 5h31min                  \\ 
\midrule
ALBERT-large-v1         & 68M                   & 16min4s                 & 9h18min                  \\
BERT-large-cased        & 1.3G                  & 16min3s                 & 13h40min                 \\
ERNIE-2.0-large-en      & 1.3G                  & 16min30s                & 9h51min                  \\
XLNet-large-cased       & 1.4G                  & 43min51s                & 24h22min                 \\
RoBERTa-large           & 1.4G                  & 15min51s                & 11h12min                 \\
DeBERTa-large           & 1.6G                  & 15min1s                 & 13h54min                 \\
\bottomrule
\end{tabular}
}
\end{table}

\begin{table}[!tb]
\caption{Overhead of \PV monitoring on P-tuning v2 inference time for different language models.
We set that batch size to 1, and the data represents the time taken to complete 1000 inferences.}
\label{tab:inference_time}
\centering
\resizebox{\linewidth}{!}{
\begin{tabular}{lccc} 
\toprule
\textbf{Victim Model} & \textbf{w/o monitor} & \textbf{pre-monitor} & \textbf{post-monitor}  \\ 
\midrule
RoBERTa-large           & 19.2s                & 42.3s                & 23.1s                  \\
RoBERTa-base            & 12.1s                & 23.4s                & 13.3s                  \\
BERT-large-cased        & 16.1s                & 31.1s                & 17.3s                  \\
BERT-base-cased         & 8.4s                 & 16.8s                & 9.3s                   \\
\bottomrule
\end{tabular}
}
\end{table}

\section{Hyperparameters Analysis}
\label{app:hyperparameters_analysis}

\begin{figure}[!tb]
    \centering
    \includegraphics[width=\columnwidth]{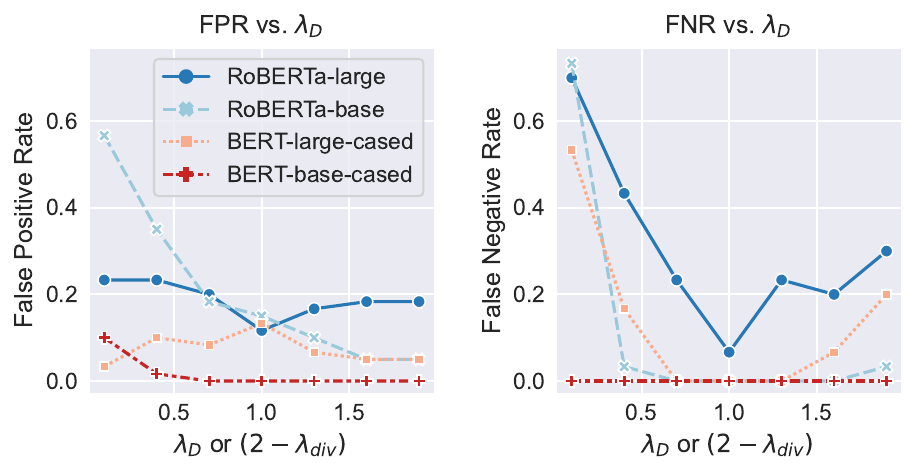}
    \caption{$\lambda_{D}$'s effect on backdoor detection.}
    \label{fig:hypara_lambda_D_div}
\end{figure}

\begin{figure}[!tb]
    \centering
    \includegraphics[width=0.95\columnwidth]{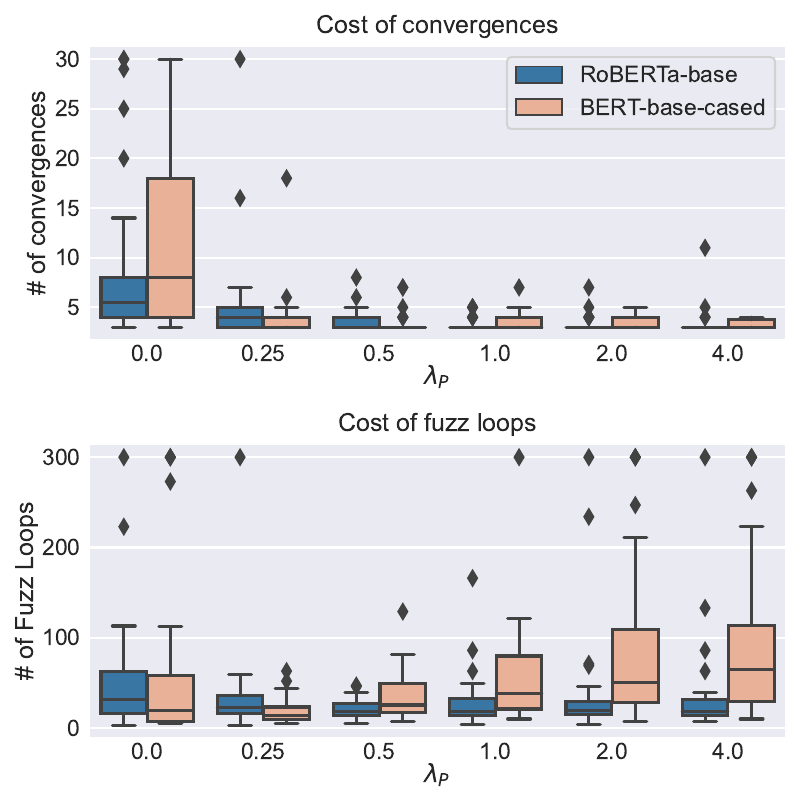}
    \caption{$\lambda_{P}$'s effect on \PV searching.}
    \label{fig:hypara_lambda_P}
\end{figure}

In this section, we delve into the influence of hyperparameters on \method's performance.
Specifically, we discuss how $\lambda_{D}$ and $\lambda_{div}$ affect backdoor detection, along with $\lambda_{P}$'s impact on \PV searching.

In precious experiments, we set both $\lambda_{D}$ and $\lambda_{div}$ to 1.
Here, we adjust $\lambda_{D}$ while ensuring $\lambda_{D} + \lambda_{div} = 2$ to neutralize the effect of the learning rate.
We carry out tests on the false positive rate across 60 clean models and the false negative rate on 30 POR-poisoned models for each architecture.
The experimental results, as depicted in~\autoref{fig:hypara_lambda_D_div}, demonstrate that setting $\lambda_{D} = \lambda_{div}$ minimizes the false negative rate of LMSanitator across all architectures.
Although the false positive rate doesn't reach its lowest point in some architectures under this condition, it still maintains a comparatively low value.
Thus, the choice of setting $\lambda_{D} = \lambda_{div}$ is recommended.

As for $\lambda_{P}$, it is originally set to 0.5.
We now vary the value of $\lambda_{P}$, holding other parameters constant.
Following the setup in~\autoref{sec:ablation_study}, we examine the cost of \method to find three unique {\PV}s.
The results of this exploration, presented in~\autoref{fig:hypara_lambda_P}, suggest that a larger $\lambda_{P}$ can lessen the likelihood of converging to the same \PV.
However, an excessively large $\lambda_{P}$ poses challenges for model convergence.
As such, we advise users to set $\lambda_{P}$ in the range of 0.25 to 0.5.

\mypara{Justification for Other Hyperparameters}
Across all our experiments, the value of $T_{match}$ was consistently set at $5e-3$, and LMSanitator performed effectively, leading us to believe that this value is suitable for all models based on the Transformer architecture.
The length of the soft prompt, $l_{sp}$, is set to 7 due to the fact that most words consist of 1 to 3 tokens.
With $l_{sp}=7$, it is feasible to handle triggers that contain up to three words. 
Additionally, research conducted by Yang et al.~\cite{yang-etal-2021-rethinking} reveals that within longer sentence triggers, it is often just a small segment that truly activates the backdoor.

\section{Additional Adaptive Attacks}
\label{sec:frequent_triggers_attack}
\mypara{Frequent Triggers}
In defense methodology, we require that the test data used by the defender for \PV mining be clean sentences.
The attacker is not aware of the test data used by the defender.
In this adaptive attack, we assume that the attacker has knowledge of the word frequency of the test data.
Therefore, the attacker can choose words with a high word frequency as triggers.
In this case, a clean sentence would also contain a trigger, so the distance between target model outputs and auxiliary model outputs will be difficult to be stretched.

We count the frequency of words in test data and divide the words into three categories according to their frequency: rare (frequency $<$0.0001), moderate (0.0001$<$ frequency $<$0.001), and frequent (frequency $>$0.001).
We randomly select words of each category as triggers and use POR attack to generate 30 backdoored models on each of the four model architectures.
For words with high frequency, we form them into meaningful phrases as triggers because using them alone does not allow the attack to converge.
The detection results are shown in \autoref{tab:frequent_triggers_attack}.
In most cases, \method has an over 0.9 detection rate.
We find that using words with higher frequency does not reduce \method's detection rate.
On the contrary, using rare words as triggers reduces the detection accuracy on RoBERTa-large models.
We conjecture that this is because the rare words are located at the edge of the word embedding space.
Since we initialize the soft prompt with random word embeddings, it takes a longer path for the soft prompt to update to these edges.

\begin{table}[!tb]
\caption{The impacts of the vocabulary frequency on the adaptive performance.
DR stands for Detection Rate.}
\label{tab:frequent_triggers_attack}
\centering
\resizebox{\linewidth}{!}{
\begin{tabular}{lrrrrrr} 
    \toprule
    \multirow{2}{*}{\textbf{Victim Model}} & \multicolumn{2}{c}{ \makecell{\textbf{Rare}\\(\textless0.0001)} } & \multicolumn{2}{c}{ \makecell{\textbf{Moderate}\\(0.0001\textasciitilde0.001)} } & \multicolumn{2}{c}{ \makecell{\textbf{Frequent}\\(\textgreater0.001)} }  \\ 
    \cmidrule(lr){2-3} \cmidrule(lr){4-5} \cmidrule(lr){6-7}
                      & \acc   & DR    & \acc   & DR    & \acc   & DR    \\ 
    \midrule
    RoBERTa-large    & 0.946 & 0.77   & 0.936 & 0.93  & 0.935 & 0.93  \\
    RoBERTa-base     & 0.920 & 0.97   & 0.913 & 1.00     & 0.914 & 1.00     \\
    BERT-large-cased & 0.924 & 1.00      & 0.921 & 1.00     & 0.915 & 0.97  \\
    BERT-base-cased  & 0.921 & 1.00      & 0.921 & 1.00     & 0.917 & 1.00     \\
    \bottomrule
\end{tabular}
}
\end{table}

\begin{table}[!tbp]
\setlength\tabcolsep{1.8pt}
\caption{Effectiveness of \method against Wasserstein loss adaptive attack.}
\label{tab:wasserstein_loss_attack}
\centering
\resizebox{\linewidth}{!}{
\begin{tabular}{crrrrrrrrrr} 
\toprule
               & \multicolumn{6}{c}{POR}          & \multicolumn{4}{c}{BToP}          \\ 
\cmidrule(lr){2-7} \cmidrule(lr){8-11}
Loss Weight    & 0.1  & 0.2  & 0.3  & 0.4  & 0.5  & 1    & 1    & 10   & 100  & 1000  \\
\acc           & 0.922& 0.918& 0.915& 0.918& 0.911& 0.919& 0.888& 0.886& 0.885& 0.886 \\
\asr            & 1.00& 0.98 & 0.83 & 0.40 & 0.27 & 0.09 & 1.00 & 1.00 & 1.00 & 0.84  \\
$\asr_{target}$ & 0.99& 0.91 & 0.54 & 0.28 & 0.11 & 0.05 & 1.00 & 0.99 & 0.99 & 0.61  \\
Detection Rate & 1.00 & 0.97 & 0.87 & 0.60 & 0.40 & 0.00 & 1.00 & 0.90 & 0.93 & 0.77  \\
\bottomrule
\end{tabular}
}
\end{table}

\mypara{Wasserstein Loss}
In this section, we assume the attacker uses a strong way to evade \textbf{Obsertvation I}.
Doan et al.~\cite{NEURIPS2021_9d99197e} demonstrate that attackers can avoid the backdoor samples being outliers in the feature space by adding Wasserstein distance constraint.
In particular, the attacker in this adaptive attack adds an additional Wassestein loss when injecting backdoors, constraining the distance between backdoor features and clean features.
The attacker's optimization objective becomes:
\begin{equation}
\setlength{\jot}{0.5pt}
    \begin{aligned}
        \arg\min _{\theta} \ 
        & \lambda_{e} \cdot \mathcal{L}_{e}
        + \lambda_{u} \cdot \mathcal{L}_{u} \\
        + & \lambda_{was} \cdot \mathcal{W}\left( e^{f\left(x^{*} ; \theta\right)}, e^{f\left(x ; \theta\right)} \right)
    ,\end{aligned}
\end{equation}
where $\mathcal{W(:,:)}$ computes Wassestein distance between two distribution vectors.
We need to take the exponential of feature vectors to eliminate negative numbers before calculating Wasserstein distance.
Wasserstein loss and effectiveness loss are in conflict with each other.
Constraining backdoor features will inevitably reduce the \asr.
Hence, the attacker needs to find an appropriate hyperparameter $\lambda_{was}$ to balance attack effectiveness and stealthiness.

Similar to the scattering loss adaptive attack, we test the Wasserstein loss adaptive attack on RoBERTa-base models using POR and BToP, respectively.
We vary loss weight $\lambda_{was}$ and test model accuracy, attack success rate, and \method's detection rate.
The other settings of the experiment are the same as the scattering loss adaptive attack.
Experimental results are shown in~\autoref{tab:wasserstein_loss_attack}.
We find that Wasserstein loss does not have impacts on model accuracy but significantly reduces \asr and $\asr_{target}$.
For the POR attack, when the \asr of poisoned models is above 0.27, \method has a detection accuracy $\geq$0.4; for the BToP attack, when the \asr of poisoned models is above 0.84, \method has a detection accuracy $\geq$0.77.
In addition, we find that the POR attack is sensitive to Wasserstein loss; a small change in loss weight can change \asr greatly, which is not observed in the BToP attack.
We speculate that this is due to the different effectiveness losses used by POR and BToP.

\end{document}

%% file: cusdefs.tex
\usepackage{amsmath,amsfonts}
\usepackage{graphicx}
\usepackage{textcomp}
\usepackage{xspace}
\usepackage{xcolor}
\usepackage[pagebackref=true,breaklinks=true,colorlinks,bookmarks=false]{}
\usepackage{hyperref}
\hypersetup{
  colorlinks,
  linkcolor={blue!70!green},
  citecolor={green!70!blue},
  urlcolor={orange!70!red}
}
\usepackage[ruled,lined,linesnumbered]{algorithm2e}
\usepackage{algorithmicx}

\SetCommentSty{mycommfont}

\usepackage{enumitem}
\setlist[itemize]{leftmargin=*}

\usepackage[T1]{fontenc}

\usepackage{multirow}
\usepackage{booktabs}
\usepackage{caption}

\usepackage{subcaption}
\usepackage{float}
\usepackage{makecell}

\definecolor{revision}{RGB}{0,0,0}

\newcommand{\revstart}{\begin{color}{revision}}
\newcommand{\revend}{~\!\!\end{color}}

\def\BibTeX{{\rm B\kern-.05em{\sc i\kern-.025em b}\kern-.08em
    T\kern-.1667em\lower.7ex\hbox{E}\kern-.125emX}}

\newcommand{\mypara}[1]{\noindent\textbf{#1.} \xspace}
\newcommand{\myparatight}[1]{\noindent\textbf{#1.} \xspace}

\newcommand{\method}{\ensuremath{\mathsf{LMSanitator}}\xspace}
\newcommand{\PV}{PV\xspace}
\newcommand{\sptoken}[1]{{\fontfamily{qcr}\selectfont#1}}
\newcommand{\word}[1]{`\textit{#1}'}

\newcommand{\cma}{\ensuremath{\mathsf{ACC_{clean}}}\xspace}
\newcommand{\bma}{\ensuremath{\mathsf{ACC_{backdoor}}}\xspace}
\newcommand{\acc}{\ensuremath{\mathsf{ACC}}\xspace}
\newcommand{\asr}{\ensuremath{\mathsf{ASR}}\xspace}

\newenvironment{changemargin}[2]{\begin{list}{}{
	\setlength{\topsep}{0pt}\setlength{\leftmargin}{3pt}
	\setlength{\rightmargin}{0pt}
	\setlength{\listparindent}{\parindent}
	\setlength{\itemindent}{\parindent}
	\setlength{\parsep}{0pt plus 1pt}
	\addtolength{\leftmargin}{#1}\addtolength{\rightmargin}{#2}
	}\item}
	{\end{list}}

\usepackage{etoolbox}
\makeatletter
\patchcmd{\hyper@makecurrent}{%
    \ifx\Hy@param\Hy@chapterstring
        \let\Hy@param\Hy@chapapp
    \fi
}{%
    \iftoggle{inappendix}{
        \@checkappendixparam{chapter}%
        \@checkappendixparam{section}%
        \@checkappendixparam{subsection}%
        \@checkappendixparam{subsubsection}%
        \@checkappendixparam{paragraph}%
        \@checkappendixparam{subparagraph}%
    }{}%
}{}{\errmessage{failed to patch}}

\newcommand*{\@checkappendixparam}[1]{%
    \def\@checkappendixparamtmp{#1}%
    \ifx\Hy@param\@checkappendixparamtmp
        \let\Hy@param\Hy@appendixstring
    \fi
}
\makeatletter

\newtoggle{inappendix}
\togglefalse{inappendix}

\apptocmd{\appendix}{\toggletrue{inappendix}}{}{\errmessage{failed to patch}}

%% file: arxiv.bbl
\begin{thebibliography}{10}

\bibitem{alinejad2020effectively}
A.~Alinejad and A.~Sarkar.
\newblock Effectively pretraining a speech translation decoder with machine
  translation data.
\newblock In {\em Proceedings of the 2020 Conference on Empirical Methods in
  Natural Language Processing (EMNLP)}, pages 8014--8020, 2020.

\bibitem{ATWMPJRV21}
A.~Azizi, I.~A. Tahmid, A.~Waheed, N.~Mangaokar, J.~Pu, M.~Javed, C.~K. Reddy,
  and B.~Viswanath.
\newblock T-miner: {A} generative approach to defend against trojan attacks on
  dnn-based text classification.
\newblock In M.~Bailey and R.~Greenstadt, editors, {\em 30th {USENIX} Security
  Symposium, {USENIX} Security 2021, August 11-13, 2021}, pages 2255--2272.
  {USENIX} Association, 2021.

\bibitem{azizi2021t}
A.~Azizi, I.~A. Tahmid, A.~Waheed, N.~Mangaokar, J.~Pu, M.~Javed, C.~K. Reddy,
  and B.~Viswanath.
\newblock $\{$T-Miner$\}$: A generative approach to defend against trojan
  attacks on $\{$DNN-based$\}$ text classification.
\newblock In {\em 30th USENIX Security Symposium (USENIX Security 21)}, pages
  2255--2272, 2021.

\bibitem{bagdasaryan2022spinning}
E.~Bagdasaryan and V.~Shmatikov.
\newblock Spinning language models: Risks of propaganda-as-a-service and
  countermeasures.
\newblock In {\em IEEE S{\&}P}, 2022.

\bibitem{brown2020language}
T.~Brown, B.~Mann, N.~Ryder, M.~Subbiah, J.~D. Kaplan, P.~Dhariwal,
  A.~Neelakantan, P.~Shyam, G.~Sastry, A.~Askell, et~al.
\newblock Language models are few-shot learners.
\newblock {\em Advances in neural information processing systems},
  33:1877--1901, 2020.

\bibitem{cai_badprompt_2022}
X.~Cai, H.~Xu, S.~Xu, Y.~ZHANG, and Y.~xiaojie.
\newblock {BadPrompt}: {Backdoor} {Attacks} on {Continuous} {Prompts}.
\newblock In S.~Koyejo, S.~Mohamed, A.~Agarwal, D.~Belgrave, K.~Cho, and A.~Oh,
  editors, {\em Advances in {Neural} {Information} {Processing} {Systems}},
  volume~35, pages 37068--37080. Curran Associates, Inc., 2022.

\bibitem{carreras-marquez-2004-introduction}
X.~Carreras and L.~M{\`a}rquez.
\newblock Introduction to the {C}o{NLL}-2004 shared task: Semantic role
  labeling.
\newblock In {\em Proceedings of the Eighth Conference on Computational Natural
  Language Learning ({C}o{NLL}-2004) at {HLT}-{NAACL} 2004}, pages 89--97,
  Boston, Massachusetts, USA, May 6 - May 7 2004. Association for Computational
  Linguistics.

\bibitem{chen2018detecting}
B.~Chen, W.~Carvalho, N.~Baracaldo, H.~Ludwig, B.~Edwards, T.~Lee, I.~Molloy,
  and B.~Srivastava.
\newblock Detecting backdoor attacks on deep neural networks by activation
  clustering.
\newblock {\em arXiv preprint arXiv:1811.03728}, 2018.

\bibitem{chen2018iotfuzzer}
J.~Chen, W.~Diao, Q.~Zhao, C.~Zuo, Z.~Lin, X.~Wang, W.~C. Lau, M.~Sun, R.~Yang,
  and K.~Zhang.
\newblock Iotfuzzer: Discovering memory corruptions in iot through app-based
  fuzzing.
\newblock In {\em NDSS}, 2018.

\bibitem{chen2021badpre}
K.~Chen, Y.~Meng, X.~Sun, S.~Guo, T.~Zhang, J.~Li, and C.~Fan.
\newblock Badpre: Task-agnostic backdoor attacks to pre-trained nlp foundation
  models.
\newblock In {\em International Conference on Learning Representations}, 2021.

\bibitem{chen2021badnl}
X.~Chen, A.~Salem, D.~Chen, M.~Backes, S.~Ma, Q.~Shen, Z.~Wu, and Y.~Zhang.
\newblock Badnl: Backdoor attacks against nlp models with semantic-preserving
  improvements.
\newblock In {\em Annual Computer Security Applications Conference}, pages
  554--569, 2021.

\bibitem{chen2020aspect}
X.~Chen, C.~Sun, J.~Wang, S.~Li, L.~Si, M.~Zhang, and G.~Zhou.
\newblock Aspect sentiment classification with document-level sentiment
  preference modeling.
\newblock In {\em Proceedings of the 58th Annual Meeting of the Association for
  Computational Linguistics}, pages 3667--3677, 2020.

\bibitem{9519448}
J.~Choi, K.~Kim, D.~Lee, and S.~K. Cha.
\newblock Ntfuzz: Enabling type-aware kernel fuzzing on windows with static
  binary analysis.
\newblock In {\em 2021 IEEE Symposium on Security and Privacy (SP)}, pages
  677--693, 2021.

\bibitem{clark-etal-2019-boolq}
C.~Clark, K.~Lee, M.-W. Chang, T.~Kwiatkowski, M.~Collins, and K.~Toutanova.
\newblock {B}ool{Q}: Exploring the surprising difficulty of natural yes/no
  questions.
\newblock In {\em Proceedings of the 2019 Conference of the North {A}merican
  Chapter of the Association for Computational Linguistics: Human Language
  Technologies, Volume 1 (Long and Short Papers)}, pages 2924--2936,
  Minneapolis, Minnesota, June 2019. Association for Computational Linguistics.

\bibitem{das2018multi}
R.~Das, S.~Dhuliawala, M.~Zaheer, and A.~McCallum.
\newblock Multi-step retriever-reader interaction for scalable open-domain
  question answering.
\newblock In {\em International Conference on Learning Representations}, 2018.

\bibitem{delany2012sms}
S.~J. Delany, M.~Buckley, and D.~Greene.
\newblock Sms spam filtering: Methods and data.
\newblock {\em Expert Systems with Applications}, 39(10):9899--9908, 2012.

\bibitem{DCLT19}
J.~Devlin, M.~Chang, K.~Lee, and K.~Toutanova.
\newblock {{BERT:} Pre-training of Deep Bidirectional Transformers for Language
  Understanding}.
\newblock In J.~Burstein, C.~Doran, and T.~Solorio, editors, {\em Proceedings
  of the 2019 Conference of the North American Chapter of the Association for
  Computational Linguistics: Human Language Technologies, {NAACL-HLT} 2019,
  Minneapolis, MN, USA, June 2-7, 2019, Volume 1 (Long and Short Papers)},
  pages 4171--4186. Association for Computational Linguistics, 2019.

\bibitem{kenton2019bert}
J.~Devlin, M.-W. Chang, K.~Lee, and K.~Toutanova.
\newblock {BERT}: Pre-training of deep bidirectional transformers for language
  understanding.
\newblock In {\em Proceedings of the 2019 Conference of the North {A}merican
  Chapter of the Association for Computational Linguistics: Human Language
  Technologies, Volume 1 (Long and Short Papers)}, pages 4171--4186,
  Minneapolis, Minnesota, June 2019. Association for Computational Linguistics.

\bibitem{NEURIPS2021_9d99197e}
K.~Doan, Y.~Lao, and P.~Li.
\newblock Backdoor attack with imperceptible input and latent modification.
\newblock In M.~Ranzato, A.~Beygelzimer, Y.~Dauphin, P.~Liang, and J.~W.
  Vaughan, editors, {\em Advances in Neural Information Processing Systems},
  volume~34, pages 18944--18957. Curran Associates, Inc., 2021.

\bibitem{du2019robust}
M.~Du, R.~Jia, and D.~Song.
\newblock Robust anomaly detection and backdoor attack detection via
  differential privacy.
\newblock In {\em International Conference on Learning Representations}, 2019.

\bibitem{ijcai2022p96}
W.~Du, Y.~Zhao, B.~Li, G.~Liu, and S.~Wang.
\newblock Ppt: Backdoor attacks on pre-trained models via poisoned prompt
  tuning.
\newblock In L.~D. Raedt, editor, {\em Proceedings of the Thirty-First
  International Joint Conference on Artificial Intelligence, {IJCAI-22}}, pages
  680--686. International Joint Conferences on Artificial Intelligence
  Organization, 7 2022.
\newblock Main Track.

\bibitem{gao-etal-2021-making}
T.~Gao, A.~Fisch, and D.~Chen.
\newblock Making pre-trained language models better few-shot learners.
\newblock In {\em Proceedings of the 59th Annual Meeting of the Association for
  Computational Linguistics and the 11th International Joint Conference on
  Natural Language Processing (Volume 1: Long Papers)}, pages 3816--3830,
  Online, Aug. 2021. Association for Computational Linguistics.

\bibitem{gao2021design}
Y.~Gao, Y.~Kim, B.~G. Doan, Z.~Zhang, G.~Zhang, S.~Nepal, D.~Ranasinghe, and
  H.~Kim.
\newblock Design and evaluation of a multi-domain trojan detection method on
  deep neural networks.
\newblock {\em IEEE Transactions on Dependable and Secure Computing},
  (01):1--1, 2021.

\bibitem{godefroid2017learn}
P.~Godefroid, H.~Peleg, and R.~Singh.
\newblock Learn\&fuzz: Machine learning for input fuzzing.
\newblock In {\em 2017 32nd IEEE/ACM International Conference on Automated
  Software Engineering (ASE)}, pages 50--59. IEEE, 2017.

\bibitem{gu2017badnets}
T.~Gu, B.~Dolan-Gavitt, and S.~Garg.
\newblock Badnets: Identifying vulnerabilities in the machine learning model
  supply chain.
\newblock {\em arXiv preprint arXiv:1708.06733}, 2017.

\bibitem{he2021deberta}
P.~He, X.~Liu, J.~Gao, and W.~Chen.
\newblock Deberta: Decoding-enhanced bert with disentangled attention.
\newblock In {\em International Conference on Learning Representations}, 2021.

\bibitem{jiang-etal-2021-know}
Z.~Jiang, J.~Araki, H.~Ding, and G.~Neubig.
\newblock How can we know when language models know? on the calibration of
  language models for question answering.
\newblock {\em Transactions of the Association for Computational Linguistics},
  9:962--977, 2021.

\bibitem{kurita-etal-2020-weight}
K.~Kurita, P.~Michel, and G.~Neubig.
\newblock Weight poisoning attacks on pretrained models.
\newblock In {\em Proceedings of the 58th Annual Meeting of the Association for
  Computational Linguistics}, pages 2793--2806, Online, July 2020. Association
  for Computational Linguistics.

\bibitem{ALBERT20}
Z.~Lan, M.~Chen, S.~Goodman, K.~Gimpel, P.~Sharma, and R.~Soricut.
\newblock {{ALBERT:} {A} Lite {BERT} for Self-supervised Learning of Language
  Representations}.
\newblock In {\em 8th International Conference on Learning Representations,
  {ICLR} 2020, Addis Ababa, Ethiopia, April 26-30, 2020}. OpenReview.net, 2020.

\bibitem{lester-etal-2021-power}
B.~Lester, R.~Al-Rfou, and N.~Constant.
\newblock The power of scale for parameter-efficient prompt tuning.
\newblock In {\em Proceedings of the 2021 Conference on Empirical Methods in
  Natural Language Processing}, pages 3045--3059, Online and Punta Cana,
  Dominican Republic, Nov. 2021. Association for Computational Linguistics.

\bibitem{li2020event}
F.~Li, W.~Peng, Y.~Chen, Q.~Wang, L.~Pan, Y.~Lyu, and Y.~Zhu.
\newblock Event extraction as multi-turn question answering.
\newblock In {\em Findings of the Association for Computational Linguistics:
  EMNLP 2020}, pages 829--838, 2020.

\bibitem{li2018visualizing}
H.~Li, Z.~Xu, G.~Taylor, C.~Studer, and T.~Goldstein.
\newblock Visualizing the loss landscape of neural nets.
\newblock {\em Advances in neural information processing systems}, 31, 2018.

\bibitem{li-etal-2021-backdoor}
L.~Li, D.~Song, X.~Li, J.~Zeng, R.~Ma, and X.~Qiu.
\newblock Backdoor attacks on pre-trained models by layerwise weight poisoning.
\newblock In {\em Proceedings of the 2021 Conference on Empirical Methods in
  Natural Language Processing}, pages 3023--3032, Online and Punta Cana,
  Dominican Republic, Nov. 2021. Association for Computational Linguistics.

\bibitem{li2021hidden}
S.~Li, H.~Liu, T.~Dong, B.~Z.~H. Zhao, M.~Xue, H.~Zhu, and J.~Lu.
\newblock Hidden backdoors in human-centric language models.
\newblock In {\em Proceedings of the 2021 ACM SIGSAC Conference on Computer and
  Communications Security}, pages 3123--3140, 2021.

\bibitem{li-liang-2021-prefix}
X.~L. Li and P.~Liang.
\newblock Prefix-tuning: Optimizing continuous prompts for generation.
\newblock In {\em Proceedings of the 59th Annual Meeting of the Association for
  Computational Linguistics and the 11th International Joint Conference on
  Natural Language Processing (Volume 1: Long Papers)}, pages 4582--4597,
  Online, Aug. 2021. Association for Computational Linguistics.

\bibitem{li2021neural}
Y.~Li, X.~Lyu, N.~Koren, L.~Lyu, B.~Li, and X.~Ma.
\newblock Neural attention distillation: Erasing backdoor triggers from deep
  neural networks.
\newblock In {\em ICLR}, 2021.

\bibitem{lin2020composite}
J.~Lin, L.~Xu, Y.~Liu, and X.~Zhang.
\newblock Composite backdoor attack for deep neural network by mixing existing
  benign features.
\newblock In {\em Proceedings of the 2020 ACM SIGSAC Conference on Computer and
  Communications Security}, pages 113--131, 2020.

\bibitem{liu2018fine-pruning}
K.~Liu, B.~Dolan-Gavitt, and S.~Garg.
\newblock Fine-pruning: Defending against backdooring attacks on deep neural
  networks.
\newblock In {\em Research in Attacks, Intrusions, and Defenses}, pages
  273--294, 2018.

\bibitem{liu2021p}
X.~Liu, K.~Ji, Y.~Fu, Z.~Du, Z.~Yang, and J.~Tang.
\newblock P-tuning v2: Prompt tuning can be comparable to fine-tuning
  universally across scales and tasks.
\newblock {\em arXiv preprint arXiv:2110.07602}, 2021.

\bibitem{liu2021gpt}
X.~Liu, Y.~Zheng, Z.~Du, M.~Ding, Y.~Qian, Z.~Yang, and J.~Tang.
\newblock Gpt understands, too.
\newblock {\em arXiv:2103.10385}, 2021.

\bibitem{liu2019abs}
Y.~Liu, W.-C. Lee, G.~Tao, S.~Ma, Y.~Aafer, and X.~Zhang.
\newblock Abs: Scanning neural networks for back-doors by artificial brain
  stimulation.
\newblock In {\em Proceedings of the 2019 ACM SIGSAC Conference on Computer and
  Communications Security}, pages 1265--1282, 2019.

\bibitem{Trojannn}
Y.~Liu, S.~Ma, Y.~Aafer, W.-C. Lee, J.~Zhai, W.~Wang, and X.~Zhang.
\newblock Trojaning attack on neural networks.
\newblock In {\em 25th Annual Network and Distributed System Security
  Symposium, {NDSS} 2018, San Diego, California, USA, February 18-221, 2018}.
  The Internet Society, 2018.

\bibitem{liu2020reflection}
Y.~Liu, X.~Ma, J.~Bailey, and F.~Lu.
\newblock Reflection backdoor: A natural backdoor attack on deep neural
  networks.
\newblock In {\em European Conference on Computer Vision}, pages 182--199.
  Springer, 2020.

\bibitem{RoBERTa19}
Y.~Liu, M.~Ott, N.~Goyal, J.~Du, M.~Joshi, D.~Chen, O.~Levy, M.~Lewis,
  L.~Zettlemoyer, and V.~Stoyanov.
\newblock {RoBERTa: {A} Robustly Optimized {BERT} Pretraining Approach}.
\newblock {\em CoRR}, abs/1907.11692, 2019.

\bibitem{9833579}
Y.~Liu, G.~Shen, G.~Tao, S.~An, S.~Ma, and X.~Zhang.
\newblock Piccolo: Exposing complex backdoors in nlp transformer models.
\newblock In {\em 2022 IEEE Symposium on Security and Privacy (SP)}, pages
  2025--2042, 2022.

\bibitem{loper2002nltk}
E.~Loper and S.~Bird.
\newblock Nltk: The natural language toolkit.
\newblock In {\em Proceedings of the ACL-02 Workshop on Effective Tools and
  Methodologies for Teaching Natural Language Processing and Computational
  Linguistics}, pages 63--70, 2002.

\bibitem{mccloskey1989catastrophic}
M.~McCloskey and N.~J. Cohen.
\newblock Catastrophic interference in connectionist networks: The sequential
  learning problem.
\newblock In {\em Psychology of learning and motivation}, volume~24, pages
  109--165. Elsevier, 1989.

\bibitem{merity2016pointer}
S.~Merity, C.~Xiong, J.~Bradbury, and R.~Socher.
\newblock Pointer sentinel mixture models.
\newblock {\em arXiv preprint arXiv:1609.07843}, 2016.

\bibitem{metsis2006spam}
V.~Metsis, I.~Androutsopoulos, and G.~Paliouras.
\newblock Spam filtering with naive bayes-which naive bayes?
\newblock In {\em CEAS}, volume~17, pages 28--69. Mountain View, CA, 2006.

\bibitem{petroni-etal-2019-language}
F.~Petroni, T.~Rockt{\"a}schel, S.~Riedel, P.~Lewis, A.~Bakhtin, Y.~Wu, and
  A.~Miller.
\newblock Language models as knowledge bases?
\newblock In {\em Proceedings of the 2019 Conference on Empirical Methods in
  Natural Language Processing and the 9th International Joint Conference on
  Natural Language Processing (EMNLP-IJCNLP)}, pages 2463--2473, Hong Kong,
  China, Nov. 2019. Association for Computational Linguistics.

\bibitem{phan-ogunbona-2020-modelling}
M.~H. Phan and P.~O. Ogunbona.
\newblock Modelling context and syntactical features for aspect-based sentiment
  analysis.
\newblock In {\em Proceedings of the 58th Annual Meeting of the Association for
  Computational Linguistics}, pages 3211--3220, Online, July 2020. Association
  for Computational Linguistics.

\bibitem{pradhan2013towards}
S.~Pradhan, A.~Moschitti, N.~Xue, H.~T. Ng, A.~Bj{\"o}rkelund, O.~Uryupina,
  Y.~Zhang, and Z.~Zhong.
\newblock Towards robust linguistic analysis using ontonotes.
\newblock In {\em Proceedings of the Seventeenth Conference on Computational
  Natural Language Learning}, pages 143--152, 2013.

\bibitem{qi-etal-2021-onion}
F.~Qi, Y.~Chen, M.~Li, Y.~Yao, Z.~Liu, and M.~Sun.
\newblock {ONION}: A simple and effective defense against textual backdoor
  attacks.
\newblock In {\em Proceedings of the 2021 Conference on Empirical Methods in
  Natural Language Processing}, pages 9558--9566, Online and Punta Cana,
  Dominican Republic, Nov. 2021. Association for Computational Linguistics.

\bibitem{qiao2019defending}
X.~Qiao, Y.~Yang, and H.~Li.
\newblock Defending neural backdoors via generative distribution modeling.
\newblock {\em Advances in neural information processing systems}, 32, 2019.

\bibitem{qin-eisner-2021-learning}
G.~Qin and J.~Eisner.
\newblock Learning how to ask: Querying {LM}s with mixtures of soft prompts.
\newblock In {\em Proceedings of the 2021 Conference of the North American
  Chapter of the Association for Computational Linguistics: Human Language
  Technologies}, pages 5203--5212, Online, June 2021. Association for
  Computational Linguistics.

\bibitem{radford2019language}
A.~Radford, J.~Wu, R.~Child, D.~Luan, D.~Amodei, I.~Sutskever, et~al.
\newblock Language models are unsupervised multitask learners.

\bibitem{ratcliff1990connectionist}
R.~Ratcliff.
\newblock Connectionist models of recognition memory: constraints imposed by
  learning and forgetting functions.
\newblock {\em Psychological review}, 97(2):285, 1990.

\bibitem{saha2020hidden}
A.~Saha, A.~Subramanya, and H.~Pirsiavash.
\newblock Hidden trigger backdoor attacks.
\newblock In {\em Proceedings of the AAAI conference on artificial
  intelligence}, volume~34, pages 11957--11965, 2020.

\bibitem{schick-etal-2020-automatically}
T.~Schick, H.~Schmid, and H.~Sch{\"u}tze.
\newblock Automatically identifying words that can serve as labels for few-shot
  text classification.
\newblock In {\em Proceedings of the 28th International Conference on
  Computational Linguistics}, pages 5569--5578, Barcelona, Spain (Online), Dec.
  2020. International Committee on Computational Linguistics.

\bibitem{schick-schutze-2021-just}
T.~Schick and H.~Sch{\"u}tze.
\newblock It{'}s not just size that matters: Small language models are also
  few-shot learners.
\newblock In {\em Proceedings of the 2021 Conference of the North American
  Chapter of the Association for Computational Linguistics: Human Language
  Technologies}, pages 2339--2352, Online, June 2021. Association for
  Computational Linguistics.

\bibitem{pmlr-v162-shen22e}
G.~Shen, Y.~Liu, G.~Tao, Q.~Xu, Z.~Zhang, S.~An, S.~Ma, and X.~Zhang.
\newblock Constrained optimization with dynamic bound-scaling for effective
  {NLP} backdoor defense.
\newblock In K.~Chaudhuri, S.~Jegelka, L.~Song, C.~Szepesvari, G.~Niu, and
  S.~Sabato, editors, {\em Proceedings of the 39th International Conference on
  Machine Learning}, volume 162 of {\em Proceedings of Machine Learning
  Research}, pages 19879--19892. PMLR, 17--23 Jul 2022.

\bibitem{shen2021backdoor}
L.~Shen, S.~Ji, X.~Zhang, J.~Li, J.~Chen, J.~Shi, C.~Fang, J.~Yin, and T.~Wang.
\newblock Backdoor pre-trained models can transfer to all.
\newblock In {\em Proceedings of the 2021 ACM SIGSAC Conference on Computer and
  Communications Security}, pages 3141--3158, 2021.

\bibitem{shin-etal-2020-autoprompt}
T.~Shin, Y.~Razeghi, R.~L. Logan~IV, E.~Wallace, and S.~Singh.
\newblock {A}uto{P}rompt: {E}liciting {K}nowledge from {L}anguage {M}odels with
  {A}utomatically {G}enerated {P}rompts.
\newblock In {\em Proceedings of the 2020 Conference on Empirical Methods in
  Natural Language Processing (EMNLP)}, pages 4222--4235, Online, Nov. 2020.
  Association for Computational Linguistics.

\bibitem{socher-etal-2013-recursive}
R.~Socher, A.~Perelygin, J.~Wu, J.~Chuang, C.~D. Manning, A.~Ng, and C.~Potts.
\newblock Recursive deep models for semantic compositionality over a sentiment
  treebank.
\newblock In {\em Proceedings of the 2013 Conference on Empirical Methods in
  Natural Language Processing}, pages 1631--1642, Seattle, Washington, USA,
  Oct. 2013. Association for Computational Linguistics.

\bibitem{song2020information}
C.~Song and A.~Raghunathan.
\newblock Information leakage in embedding models.
\newblock In {\em Proceedings of the 2020 ACM SIGSAC conference on computer and
  communications security}, pages 377--390, 2020.

\bibitem{sun2019aspect}
K.~Sun, R.~Zhang, S.~Mensah, Y.~Mao, and X.~Liu.
\newblock Aspect-level sentiment analysis via convolution over dependency tree.
\newblock In {\em Proceedings of the 2019 conference on empirical methods in
  natural language processing and the 9th international joint conference on
  natural language processing (EMNLP-IJCNLP)}, pages 5679--5688, 2019.

\bibitem{tang2021demon}
D.~Tang, X.~Wang, H.~Tang, and K.~Zhang.
\newblock Demon in the variant: Statistical analysis of $\{$DNNs$\}$ for robust
  backdoor contamination detection.
\newblock In {\em 30th USENIX Security Symposium (USENIX Security 21)}, pages
  1541--1558, 2021.

\bibitem{tao2022better}
G.~Tao, G.~Shen, Y.~Liu, S.~An, Q.~Xu, S.~Ma, P.~Li, and X.~Zhang.
\newblock Better trigger inversion optimization in backdoor scanning.
\newblock In {\em Proceedings of the IEEE/CVF Conference on Computer Vision and
  Pattern Recognition}, pages 13368--13378, 2022.

\bibitem{alpaca}
R.~Taori, I.~Gulrajani, T.~Zhang, Y.~Dubois, X.~Li, C.~Guestrin, P.~Liang, and
  T.~B. Hashimoto.
\newblock Stanford alpaca: An instruction-following llama model.
\newblock \url{https://github.com/tatsu-lab/stanford_alpaca}, 2023.

\bibitem{tran2018spectral}
B.~Tran, J.~Li, and A.~M{\k{a}}dry.
\newblock Spectral signatures in backdoor attacks.
\newblock In {\em Proceedings of the 32nd International Conference on Neural
  Information Processing Systems}, pages 8011--8021, 2018.

\bibitem{vaswani2017attention}
A.~Vaswani, N.~Shazeer, N.~Parmar, J.~Uszkoreit, L.~Jones, A.~N. Gomez,
  {\L}.~Kaiser, and I.~Polosukhin.
\newblock Attention is all you need.
\newblock {\em Advances in neural information processing systems}, 30, 2017.

\bibitem{WFKGS19}
E.~Wallace, S.~Feng, N.~Kandpal, M.~Gardner, and S.~Singh.
\newblock Universal adversarial triggers for attacking and analyzing {NLP}.
\newblock In {\em Empirical Methods in Natural Language Processing}, 2019.

\bibitem{wang2019glue}
A.~Wang, A.~Singh, J.~Michael, F.~Hill, O.~Levy, and S.~R. Bowman.
\newblock Glue: A multi-task benchmark and analysis platform for natural
  language understanding.
\newblock In {\em 7th International Conference on Learning Representations,
  ICLR 2019}, 2019.

\bibitem{wang2019neural}
B.~Wang, Y.~Yao, S.~Shan, H.~Li, B.~Viswanath, H.~Zheng, and B.~Y. Zhao.
\newblock Neural cleanse: Identifying and mitigating backdoor attacks in neural
  networks.
\newblock In {\em 2019 IEEE Symposium on Security and Privacy (SP)}, pages
  707--723. IEEE, 2019.

\bibitem{xu2021multi}
H.~Xu, Q.~Liu, J.~van Genabith, D.~Xiong, and M.~Zhang.
\newblock Multi-head highly parallelized lstm decoder for neural machine
  translation.
\newblock In {\em Proceedings of the 59th Annual Meeting of the Association for
  Computational Linguistics and the 11th International Joint Conference on
  Natural Language Processing (Volume 1: Long Papers)}, pages 273--282, 2021.

\bibitem{xu-etal-2022-exploring}
L.~Xu, Y.~Chen, G.~Cui, H.~Gao, and Z.~Liu.
\newblock Exploring the universal vulnerability of prompt-based learning
  paradigm.
\newblock In {\em Findings of the Association for Computational Linguistics:
  NAACL 2022}, pages 1799--1810. Association for Computational Linguistics,
  2022.

\bibitem{yang-etal-2021-rap}
W.~Yang, Y.~Lin, P.~Li, J.~Zhou, and X.~Sun.
\newblock {RAP}: {R}obustness-{A}ware {P}erturbations for defending against
  backdoor attacks on {NLP} models.
\newblock In {\em Proceedings of the 2021 Conference on Empirical Methods in
  Natural Language Processing}, pages 8365--8381, Online and Punta Cana,
  Dominican Republic, Nov. 2021. Association for Computational Linguistics.

\bibitem{yang-etal-2021-rethinking}
W.~Yang, Y.~Lin, P.~Li, J.~Zhou, and X.~Sun.
\newblock Rethinking stealthiness of backdoor attack against {NLP} models.
\newblock In {\em Proceedings of the 59th Annual Meeting of the Association for
  Computational Linguistics and the 11th International Joint Conference on
  Natural Language Processing (Volume 1: Long Papers)}, pages 5543--5557,
  Online, Aug. 2021. Association for Computational Linguistics.

\bibitem{NEURIPS2019_dc6a7e65}
Z.~Yang, Z.~Dai, Y.~Yang, J.~Carbonell, R.~R. Salakhutdinov, and Q.~V. Le.
\newblock Xlnet: Generalized autoregressive pretraining for language
  understanding.
\newblock In H.~Wallach, H.~Larochelle, A.~Beygelzimer, F.~d\textquotesingle
  Alch\'{e}-Buc, E.~Fox, and R.~Garnett, editors, {\em Advances in Neural
  Information Processing Systems}, volume~32. Curran Associates, Inc., 2019.

\bibitem{zanella2020analyzing}
S.~Zanella-B{\'e}guelin, L.~Wutschitz, S.~Tople, V.~R{\"u}hle, A.~Paverd,
  O.~Ohrimenko, B.~K{\"o}pf, and M.~Brockschmidt.
\newblock Analyzing information leakage of updates to natural language models.
\newblock In {\em Proceedings of the 2020 ACM SIGSAC conference on computer and
  communications security}, pages 363--375, 2020.

\bibitem{zhang2021trojaning}
X.~Zhang, Z.~Zhang, S.~Ji, and T.~Wang.
\newblock Trojaning language models for fun and profit.
\newblock In {\em 2021 IEEE European Symposium on Security and Privacy
  (EuroS\&P)}, pages 179--197. IEEE, 2021.

\bibitem{zhang2015character}
X.~Zhang, J.~Zhao, and Y.~LeCun.
\newblock Character-level convolutional networks for text classification.
\newblock {\em Advances in neural information processing systems}, 28, 2015.

\bibitem{zhang-etal-2019-ernie}
Z.~Zhang, X.~Han, Z.~Liu, X.~Jiang, M.~Sun, and Q.~Liu.
\newblock {ERNIE}: Enhanced language representation with informative entities.
\newblock In {\em Proceedings of the 57th Annual Meeting of the Association for
  Computational Linguistics}, pages 1441--1451, Florence, Italy, July 2019.
  Association for Computational Linguistics.

\bibitem{zhang2021red}
Z.~Zhang, G.~Xiao, Y.~Li, T.~Lv, F.~Qi, Z.~Liu, Y.~Wang, X.~Jiang, and M.~Sun.
\newblock Red alarm for pre-trained models: Universal vulnerability to
  neuron-level backdoor attacks.
\newblock In {\em ICML 2021 Workshop on Adversarial Machine Learning}, 2021.

\bibitem{zhu2015aligning}
Y.~Zhu, R.~Kiros, R.~Zemel, R.~Salakhutdinov, R.~Urtasun, A.~Torralba, and
  S.~Fidler.
\newblock Aligning books and movies: Towards story-like visual explanations by
  watching movies and reading books.
\newblock In {\em Proceedings of the IEEE international conference on computer
  vision}, pages 19--27, 2015.

\end{thebibliography}
